\documentclass[lettersize,journal]{IEEEtran}
\usepackage{amsmath,amsfonts,amssymb}
\usepackage{algorithmic}
\usepackage{algorithm}
\usepackage{array}
\usepackage[caption=false,font=normalsize,labelfont=sf,textfont=sf]{subfig}
\usepackage{textcomp}
\usepackage{stfloats}
\usepackage{url}
\usepackage{verbatim}
\usepackage{graphicx}
\usepackage{cite}
\usepackage{color}

\usepackage{multirow}
\usepackage{makecell}

\usepackage{url}

\usepackage[breaklinks]{hyperref}
\usepackage{breakurl}

\usepackage[absolute]{textpos}

\usepackage{balance}
\usepackage{verbatim}

\begin{document}


\title{Lattice-based shape tracking and servoing \\ of elastic objects}

\author{Mohammadreza Shetab-Bushehri, Miguel Aranda, Youcef Mezouar and Erol \"{O}zg\"{u}r

\thanks{This work was supported by project SOFTMANBOT, which
received funding from the European Union's Horizon 2020 research
and innovation programme under grant agreement No 869855.}
\thanks{This work was also supported by \mbox{MCIN/AEI/10.13039/501100011033}, the ERDF A way of making Europe, and the European Union NextGenerationEU/PRTR under Projects \mbox{PID2021-124137OB-I00} and \mbox{TED2021-130224B-I00}; and by the Spanish Ministry of Universities and the European Union-NextGenerationEU under a María Zambrano Fellowship.}
\thanks{MR. Shetab-Bushehri, Y. Mezouar and E. \"{O}zg\"{u}r are with the CNRS, Clermont Auvergne INP, Institut Pascal, Universit\'{e} Clermont Auvergne, F-63000 Clermont-Ferrand, France (e-mail: m.r.shetab@gmail.com; \hspace{3.5pt}  youcef.mezouar@sigma-clermont.fr; erolozgur@gmail.com).
M. Aranda is with the Instituto de Investigación en Ingeniería de Aragón (I3A), Universidad de Zaragoza, E-50018 Zaragoza, Spain (e-mail: miguel.aranda@unizar.es). }
}



\maketitle

\begin{abstract}
In this paper, we propose a general unified tracking-servoing approach for controlling the shape of elastic deformable objects using robotic arms. Our approach works by forming a lattice around the object, binding the object to the lattice, and tracking and servoing the lattice instead of the object. This makes our approach have full control over the deformation of elastic deformable objects of any general form (linear, thin-shell, volumetric) in 3D space. Furthermore, it decouples the runtime complexity of the approach from the objects’ geometric complexity. Our approach is based on the As-Rigid-As-Possible (ARAP) deformation model. It requires no mechanical parameter of the object to be known and can drive the object toward desired shapes through large deformations. The inputs to our approach are the point cloud of the object's surface in its rest shape and the point cloud captured by a 3D camera in each frame. 
Overall, our approach is more broadly applicable than existing approaches.
We validate the efficiency of our approach through numerous experiments with elastic deformable objects of various shapes and materials (paper, rubber, plastic, foam). Experiment videos are available on the project website: \renewcommand\UrlFont{\rmfamily\itshape}
\url{https://sites.google.com/view/tracking-servoing-approach}.

\end{abstract}

\begin{IEEEkeywords}
Visual Servoing, Visual Tracking, Sensor-based Control, Manipulation of Deformable Objects.
\end{IEEEkeywords}


\begin{textblock}{60}(.58,0.4)
\noindent \hspace{0pt} \scriptsize{\hspace{26.5pt}\textcolor{blue}{This is the author's version of an article published in IEEE Transactions on Robotics.  \\  \indent \indent \indent The publisher's version and the citation information are available at https://doi.org/10.1109/TRO.2023.3331596}}
\end{textblock}

\begin{textblock}{60}(1.35,15.50)
\noindent \hspace{0pt}\scriptsize{\textcolor{blue}{\copyright \hspace{.4pt} 2023 IEEE. Personal use is permitted, but republication/redistribution requires IEEE permission.
See https://www.ieee.org/publications/rights/index.html for more information.}}
\end{textblock}

\section{Introduction}
\IEEEPARstart{R}{obotic} manipulation of deformable objects is one of the branches of robotics that has attracted the most attention in recent years. This is because of many practical applications, including manufacturing \cite{Zhu2018dual, aranda2020monocular, koessler2021efficient, shetab2022rigid}, surgical \cite{Navarro2018,AlambeigiRAL2019,Shin2019} and household \cite{Hu2018,Hu2019}. However, existing challenges in tracking and manipulating deformable objects prevented many of these applications from being put into practice. These challenges mainly stem from the well-known natural complexities of deformable objects compared to rigid objects, i.e., high degrees of freedom, various types of deformations, occlusions, and self-occlusions. 
In the literature, manipulation of deformable objects toward a specific desired shape using sensory feedback is named shape servoing \cite{Navarro2018}. In shape servoing, elastic objects have been the subject of interest in many practical applications. This stems from the fact that many objects around us behave elastically, i.e., tend to keep their rest shape during the manipulation.
Shape servoing approaches for elastic deformable  objects have been mainly categorized based on the general form of the object being manipulated, i.e., linear \cite{koessler2021efficient, Zhu2018dual, lagneau2020automatic, qi2022contour, bretl2014quasi, sintov2020motion, lv2022dynamic, wang2022offline, aghajanzadeh2022adaptive, aghajanzadeh2022asap, aghajanzadeh2022optimal}, thin-shell \cite{berenson2013manipulation, McConachie2020, Hu2018, Zhu2021, Hu2019, shetab2022rigid}, and volumetric \cite{thach2022learning, Ficuciello2018, Zhu2021, Hu2019}.
This is due to the distinct deformation characteristics and particular assumptions and simplifications that can be made for each one of these object forms. 
Among all of presented shape servoing approaches, there are several that can be applied on two or three different forms of the object \cite{McConachie2020, Zhu2021, Hu2019}. These approaches, however, are merely tested on simple scenarios where the initial and desired shapes of the elastic deformable object are almost aligned. Indeed, the literature lacks a general shape servoing approach capable of being exploited as a universal solution for all forms of elastic deformable objects with any geometry and having full control over the object's deformation in 3D space. 

\begin{figure}[!t]\centering
    \includegraphics[trim={0 10pt 0 10pt},width=.90\linewidth]{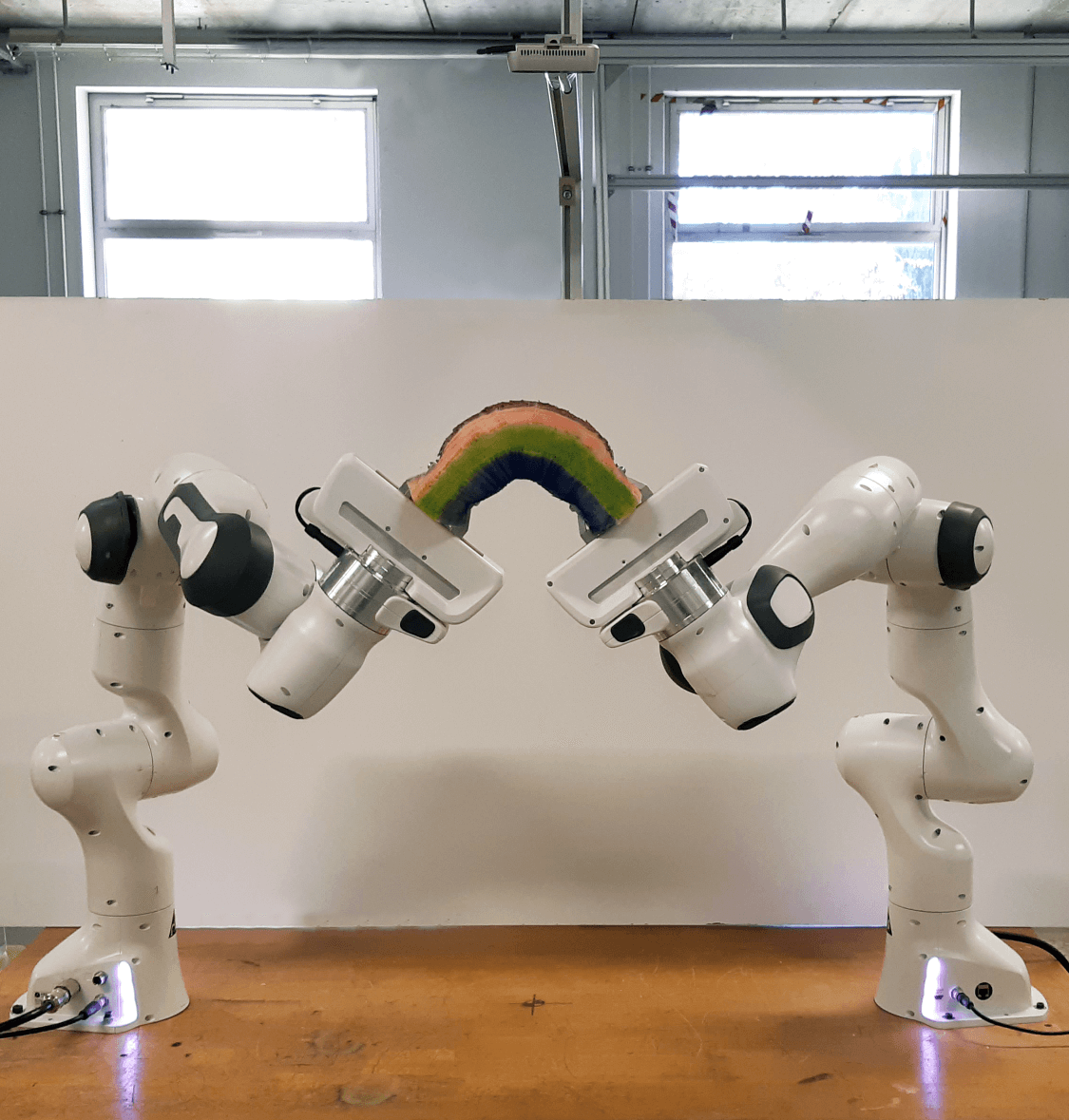} 
    \caption{Our experimental setup with two Franka Emika robots manipulating a foam octagonal cylinder. An overhead 3D camera provides the input for our approach.}
    \label{fig:setup}
\end{figure}


In this paper, we propose a novel general unified tracking-servoing approach that addresses this issue.
The approach is composed of a tracking pipeline and a servoing pipeline. The tracking pipeline tracks the shape of the elastic deformable object in 3D at every instant. The servoing pipeline employs the tracking data to drive the object to a desired shape.
Our approach is \textit{general} as it can handle linear, thin-shell and volumetric elastic objects, not being limited to a specific form of object. It is \textit{unified} as we solve jointly the two problems of tracking and servoing.
A central idea of our approach is to form a 3D lattice around the object. This lattice is bound to the object by geometrical constraints. We track and servo this lattice instead of the object itself. This is done by applying visual and deformation constraints on the lattice. We use the well-known As-Rigid-As-Possible (ARAP) model \cite{Sorkine2007} as the deformation constraint. This model is based on a geometrical representation of elastic objects' tendency for keeping their local rigidity. 
We present an analytical expression of the deformation Jacobian of the ARAP model. This deformation Jacobian, which we use on our lattice, is our model of how the robot motions deform the lattice. We obtain the Jacobian's expression by rearranging the ARAP deformation equation and exploiting modeling tools presented in \cite{sorkine2004laplacian, koessler2021efficient, aghajanzadeh2022asap}. Our Jacobian expression is directly applicable to both the whole object (full shape), or a region of it (partial shape). This gives us the useful ability to perform both full or partial shape servoing.
We propose a robotic control law based on this Jacobian. Using this control law, we servo the lattice and thus the object toward the desired shape. The inputs to our approach are the point cloud of the object's surface in its rest shape and the point cloud captured by a 3D camera in each frame. 
We use ARAP both for tracking and for servoing. ARAP has been employed frequently for tracking elastic deformable objects \cite{collins2016robust, parashar2015rigid, fuentes2022deep, fuentes2021texture}, including in robotic manipulation scenarios \cite{Han2018, Hu2019}. We first proposed to use ARAP for servoing in \cite{shetab2022rigid}; our main insight was that, when used in a shape-feedback control loop, ARAP provides a precise enough representation of the elastic object's shape dynamics. Overall, ARAP is a simple and well-established model and it allows working with elastic deformable objects without having to use mechanical parameters. In this paper, we show how to achieve both tracking and servoing requiring only this single model.

The idea of exploiting a lattice to manage the object's deformation has been previously utilized in manipulating shapes in computer graphics when the shape's representative mesh is too high-resolution to be directly manipulated \cite{zollhofer2012gpu}.
We use the same idea to transfer the object's deformation to the lattice and simplify the deformation. This is reasonable as for many objects, despite all the details on their surface (which are represented by a detailed mesh), the object cannot be deformed in more than several mostly simple ways. These deformations can, thus, be effectively represented by the lattice encapsulating the object. In fact, we solve the deformation equations for the lattice instead of the object. At the same time, we use geometrical constraints between the lattice and the object as binding constraints. This, concretely, decouples the runtime complexity of the approach from the objects’ geometric complexity and makes the whole tracking-servoing process run much faster. 
Furthermore, using a lattice makes our approach capable of being generalized for any form of the object (linear, thin-shell, volumetric). This is due to the fact that we develop our tracking-servoing approach for a 3D lattice that is formed in the same way for any object geometry. While our approach follows the standard steps in sensor-based shape control, our innovative use of a lattice and geometrical modeling allows us to improve in terms of generality, unification of tracking and servoing, efficiency and simplicity. Our specific contributions are detailed next.

\textit{Contribution}. We list our contributions as follows:
\begin{itemize}

\item We present a general unified tracking-servoing approach for elastic objects that can handle any object form (linear, thin-shell,
volumetric) with any geometry. Our approach does not require any mechanical parameter of the object to be known. Instead, we only use a point cloud that represents the object's surface with sufficient accuracy.


\item Our approach does not use the texture of the object. It only employs the point cloud captured in each frame by a 3D camera as the input. This brings the advantage to our approach that it works with objects without specific textures.

\item We present a novel analytical expression of the deformation Jacobian of the ARAP model. This avoids the need for numerical approximations.

\item Our approach has full control over the object's deformation in 3D space. This means that we can start servoing the object from an initial shape, translate, rotate, and deform the object toward a desired shape that is characterized by a totally different visible part of the object in comparison to the visible part of the object at the initial configuration. To our knowledge, this is the first approach that is proved to handle such scenarios in practice. 

\item The idea of using a lattice makes tracking and servoing much faster as we deal with the deformation equations for the lattice and not the object. This makes the execution speed of the approach to a high extent independent of the object's geometric complexities. As a result, our approach runs fast and is needless of any specific hardware. 
The execution speed of our tracking-servoing approach reaches 20-30\,FPS during our experiments without any parallelization using only CPU. 

\item The servoing pipeline can servo the whole or a part of the object. The definition of the servoed regions of the objects and its implementation in the servoing pipeline is easy and straightforward.

\item Our approach is easily scalable in terms of increasing the number of manipulating robots. Moreover, increasing the number of robots does not impose considerable additional execution time on the servoing approach.

\end{itemize}

We present a diverse experimental evaluation where our approach shows remarkable performance. Our setup is illustrated in figure \ref{fig:setup}. We use objects of different forms (linear, thin-shell, volumetric) and materials (paper, rubber, plastic, foam). Our approach can handle particularly difficult scenarios including: large non-planar deformation of a cable, servoing of a convoluted foam with a spiky surface, partial servoing of disjoint object regions, or simultaneous twisting and bending of a volumetric foam. We also illustrate that our approach has limitations in certain scenarios that require additional techniques (e.g., driving the object through its singular shape).

\section{Related work}
\subsection{Shape tracking of deformable objects} \label{subsec:Shape_tracking_of_deformable_objects} 
Tracking deformable objects has been the subject of many studies in computer vision. The main criterion for categorizing these methods is the type of used sensor, i.e., monocular camera \cite{bartoli2015shape, parashar2015rigid, chhatkuli2016stable,famouri2018fast, ozgur2017particle, aranda2020monocular, ngo2015template, collins2015poster, collins2016robust, liu2016better,pumarola2018geometry, golyanik2018hdm, shimada2019ismo, fuentes2022deep, fuentes2021texture, habermann2018nrst}, or 3D camera \cite{petit2015real, tang2018track, newcombe2015dynamicfusion, tsoli2016tracking}. Despite the considerable progress made in this area, the employment of these methods in robotic studies was limited.
This stems from several reasons: 
\textit{(i)} The codes of the deformable object tracking methods in the literature are not open access, or if they are, they are challenging to adapt to a new problem. 
\textit{(ii)} These methods are primarily object-specific, meaning that plenty of adjustments should be made to apply them to a new object. In the case of analytical methods, these adjustments might include creating a template of the object (comprising 3D mesh, texture-map, and UV map) and setting the mechanical parameters of the object \cite{famouri2018fast, collins2015poster,ngo2015template, collins2016robust, petit2015real}.
Regarding the neural-network-based methods, these adjustments include creating a new synthetic dataset of the new object of interest and training the whole network from the beginning with that new synthetic dataset, followed by a possible fine-tuning step \cite{pumarola2018geometry,golyanik2018hdm,shimada2019ismo,fuentes2022deep,fuentes2021texture}.  
\textit{(iii)} Many of the presented tracking methods in the literature are specified to a particular object form including linear \cite{tang2018track} or thin-shell \cite{bartoli2015shape,chhatkuli2016stable,famouri2018fast, habermann2018nrst, newcombe2015dynamicfusion, tsoli2016tracking}. As a result, they cannot be used as a general tracking solution to different object forms. 
\textit{(iv)} As these tracking methods are presented in the field of computer vision, they have not exploited the common priors existing in the field of robotics that can be considerably helpful throughout the tracking period, such as the known coordinates of the robotic grippers.
As a result of the mentioned reasons, researchers in the field of robotics have turned to track a simplified representation of objects instead of the whole shape. This simplified representation can be a handful of points on the surface of the object \cite{Navarro2013, Navarro2014, Ficuciello2018, Shin2019, aranda2020monocular, McConachie2020, aghajanzadeh2022asap} or the point cloud of the visible part of the object that remains almost the same throughout the manipulation \cite{Hu2018, Hu2019, thach2022learning}.

In contrast to the previous works in robotics, we present a robust shape tracking pipeline along with our shape servoing pipeline that fully tracks the shape of objects of any form and geometry. A point cloud of the object in its rest shape is the only information that our pipeline requires. It is also well designed for robotic applications as it exploits common priors such as robotic grippers' known coordinates. 
The novel idea of using a lattice to simplify object deformation places our shape tracking pipeline among the fastest methods in the literature.
This is while our tracking pipeline does not need any special hardware to run. We highlight that the shape tracking pipeline can also be used independently of our servoing pipeline.

\subsection{Model-based shape servoing} \label{subsec:model-based_Shape_servoing} 
Deformation models facilitate shape servoing of deformable objects by providing the knowledge of objects' behavior under deformation.
We categorize the deformation models as mechanical and geometrical models. Mechanical models predict the object's shape by solving mechanical equations under constraints of force or displacement. 
Finite Element Method (FEM) has been the most widely-used mechanical model in the literature \cite{Duenser2018, Ficuciello2018, koessler2021efficient, Sanchez2020}.
In this context, \cite{Duenser2018} presented an open-loop simulation-based control methodology wherein the desired deformations are directly mapped to joint angle commands.
\cite{Ficuciello2018} used FEM in an open-loop control approach for in-hand soft object manipulation. 
\cite{koessler2021efficient} introduced a closed-loop controller employing a computationally-efficient FEM that exploits a partition of mesh nodes. The authors applied their approach to a linear object. 
In \cite{Sanchez2020}, the authors used only force feedback along with FEM to servo a volumetric object.
\cite{lv2022dynamic} presented a dynamic model that simulated stretching, bending, and twisting deformation in linear objects. The authors exploited this model to deploy a linear object onto a plane using both single-arm and dual-arm setups.
Although mechanical models can be considered a reliable solution for predicting object deformation, the computation is often costly and depends on the intrinsic mechanical properties of the object which vary from one to another. 
On the contrary, our proposed shape servoing pipeline is fast and does not require any prior information on the mechanical parameters of the object.

Unlike mechanical models, geometrical models simulate the object deformation considering only geometrical constraints such as preserving the local rigidity. This makes these models independent of the mechanical parameters of the object.
\cite{aranda2020monocular} drove the object toward the desired shape by defining intermediary shapes. The intermediary shapes were calculated by applying position-based dynamics \cite{muller2007} as the deformation model.
In \cite{shetab2022rigid}, we proposed to employ the ARAP deformation model to fully servo thin-shell objects. We computed a deformation Jacobian via numerical differentiation: specifically, in simulation, we perturbed each controlled DOF and computed the resulting change of the object’s shape using ARAP. This shape servoing approach was shown to be fast and could be used for objects of various materials.
Later, \cite{Giraud2022optimal, aghajanzadeh2022optimal} exploited the same deformation Jacobian with an optimal controller.
Recently, \cite{aghajanzadeh2022asap} proposed a novel offline Jacobian to be used in servoing linear objects in 2D space. This Jacobian was based on an As-Similar-As-Possible (ASAP) deformation model, in which the object has a tendency to preserve its original shape up to a similarity
transformation of that shape.
%

In comparison to these studies which are restricted to
either linear or thin-shell objects, our proposed shape servoing
pipeline can be applied to objects of any form thanks to the
novel idea of exploiting a lattice.

\subsection{Model-free shape servoing} \label{subsec:model-free_Shape_servoing} 
Model-free shape servoing can be considered as the most studied approach in the literature. 
In these approaches, no prior information on the object deformation is required. 
We divide these approaches into two main categories: sensor-based deformation Jacobian, and geometric heuristics.
As for the first category of approaches, online sensor measurements are employed to estimate a deformation Jacobian to be used in shape servoing \cite{Navarro2013, Navarro2014, Navarro2018, Hu2018, Zhu2018dual, AlambeigiRAL2019, lagneau2020automatic, Zhu2021}.
These sensor measurements are taken by a 2D or 3D camera while the object is manipulated by robots.
This Jacobian is then used to control a simplified representation of the object, i.e., several sampled points on the object's surface \cite{Navarro2013, Navarro2014, Hu2018, AlambeigiRAL2019}, or the object's contour \cite{Navarro2018, Zhu2018dual, Zhu2021} in the image space.
In general, requiring the robot's motion for calculating the Jacobian makes
these approaches more complex, and sensitive to noise.
Conversely, our proposed approach is less noise-sensitive, we can control the full shape of the object in 3D space, and we compute our Jacobian without needing robot motions.


Works \cite{berenson2013manipulation, ruan2018accounting, McConachie2020} exploit a deformation Jacobian estimated using a geometric heuristic (diminishing rigidity). This approach is fast and has been tested in simulation and real experiments with simple deformations. However, the lack of an object deformation model in calculating the Jacobian limits its practicability in many real scenarios. In contrast to this approach, our Jacobian is calculated analytically based on a geometric deformation model.

\subsection{Learning-based shape servoing} \label{subsec:learning-based_Shape_servoing} 
In recent years, the field of deformable object manipulation has been aligned with the growing trend toward using learning-based approaches. Different studies have been conducted in this field \cite{matas2018sim, Hu2019, Shin2019, jangir2020dynamic, thach2022learning, laezza2021learning, melodie2022Robotic} among which reinforcement learning (RL) was the most widely-used approach \cite{matas2018sim, jangir2020dynamic, laezza2021learning, melodie2022Robotic}.
Despite the relative success of these approaches, they still suffer from significant limitations. 
First, they are specified to a particular type of the object such as linear \cite{laezza2021learning, melodie2022Robotic} or thin-shell \cite{matas2018sim, Shin2019, jangir2020dynamic} or just a surface of a volumetric object \cite{Hu2019, thach2022learning}.
Furthermore, they have been mostly tested in simulation, and if in real, the deformations have been simple \cite{Hu2019, Shin2019, thach2022learning}. 
Except in \cite{Hu2019} where an online learning process was used to train the agent (i.e., robot) in real, the other studies trained their agents in simulation with objects of specific mechanical parameters. This makes them suffer from the well-known sim-to-real gap. In addition, training the agent for a new object requires a lot of data to provide, which is a tedious task for deformable objects.

In contrast to these approaches, we do not need any training and our shape servoing pipeline provides full control over the deformation of elastic objects of any form.

\begin{figure*}[!t]\centering
    \includegraphics[trim={0 10pt 0 10pt},width=.90\linewidth]{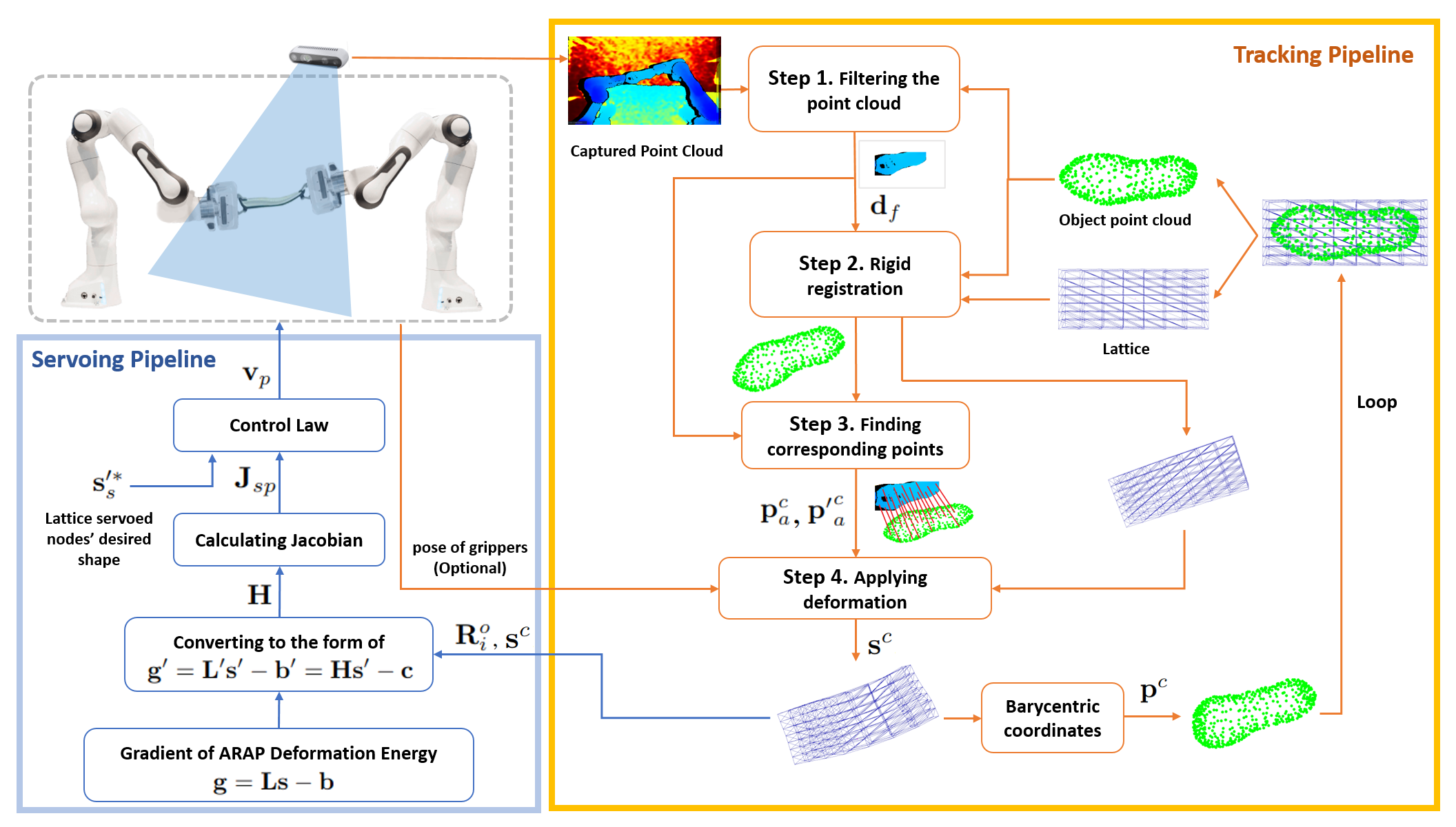} 
    \caption{  Flowchart of the proposed shape tracking-servoing approach. In each frame, the tracking pipeline takes the captured point cloud in the current frame and $\mathbf{s}$ (and thus $\mathbf{p}$) from the previous frame as the input and estimates $\mathbf{s}^{c}$ (and thus $\mathbf{p}^{c}$) in the current frame. $\mathbf{s}^{c}$ and $\mathbf{R}^{o}_i$ are sent to the servoing pipeline where an analytical expression for a deformation Jacobian is calculated. We use this Jacobian in a control law to send proper commands to the robots to guide $\mathbf{s}^{c}$ toward $\mathbf{s}^{*}$ and consequently $\mathbf{p}^{c}$ toward $\mathbf{p}^{*}$.   }
    \label{fig:flowchart}
\end{figure*}

\begin{figure} \centering
    \includegraphics[trim={0 10pt 0 10pt},width=.90\linewidth]{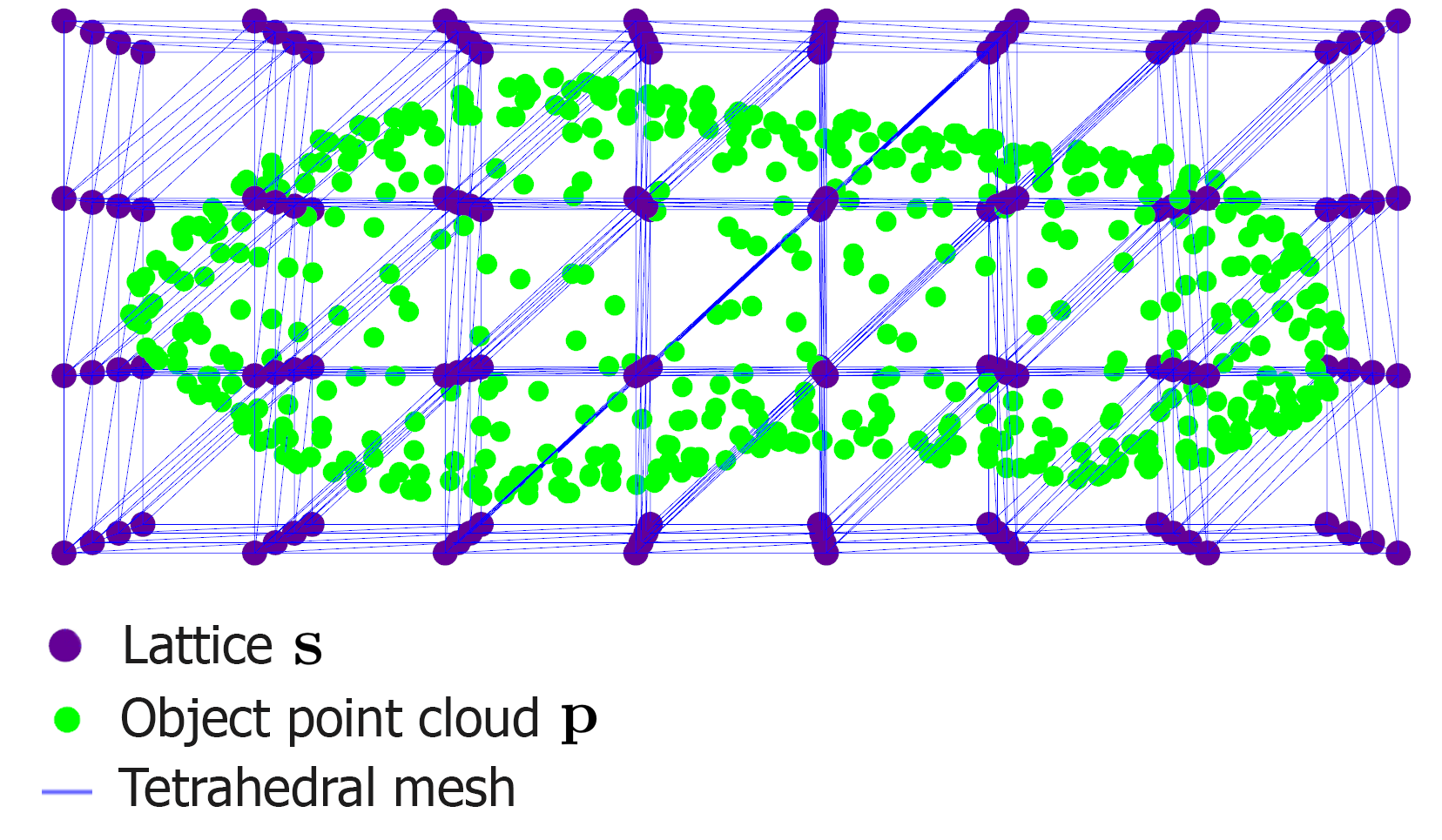} 
    \caption{Lattice $\mathbf{s}$ encapsulating the point cloud $\mathbf{p}$ of a shoe sole at its rest shape. We also generate a tetrahedral mesh over the lattice nodes.}
    \label{fig:lattice}
\end{figure}

\section{Approach description and notation} \label{sec:Problem_formulation}
The problem that we address is tracking an elastic deformable object with any general form (linear, thin-shell, and volumetric) and manipulating it by robotic arms in a feedback control loop so that its shape gradually conforms to a desired shape.
To formulate this problem, we use the following notation throughout the paper:
\begin{itemize}
\item Scalars: italic lower-case letters.
\item Vectors: bold lower-case letters.
\item Matrices: bold upper-case letters. 
\item Sets: calligraphic letters.
\end{itemize}
Note that for some developments we represent vectors of points as two-dimensional arrays (i.e., with one column for each spatial dimension), following references \cite{zollhofer2012gpu,Sorkine2007}. We define flattening a vector as transposing the rows of the vector and stacking them together in a one-column vector. For example, given a vector $\mathbf{w}$ of dimension $m\times n$, we define the flattened $\mathbf{w}$ as $\mathbf{w_f}$ of dimension $mn\times 1$.

We start by considering a set of robots $\mathcal{M} = \left \{1,2,...,m \right \}$ that firmly grasp the object during manipulation. We also assume that the forward kinematic models of the robots are known.  
We use a 3D camera as the sensor providing the input point cloud data. The relative poses between the camera and the robots are assumed to be known.
Our goal is to track the object at each instant and introduce a control scheme that computes the 6-DOF velocity vectors associated with all the robots' grippers stacked in a column vector of length $6m$: $\mathbf{v}=[\mathbf{v}^\intercal_1, ..., \mathbf{v}^\intercal_m]^\intercal$.
Figure \ref{fig:flowchart} illustrates the flowchart of our proposed approach.

We use a point cloud of the object’s surface to represent the object’s shape. This point cloud is generated when the object is undeformed, i.e., is at its rest shape. We use the $n\times 3$ vector $\mathbf{p}$ to represent this point cloud. The resolution of this point cloud should be fine enough to represent the object's geometry with sufficient accuracy throughout the tracking period. This will be explained more in Sect. \ref{sec:tracking_pipeline}. 
We, then, define the lattice $\mathbf{s}$, an $n_l\times 3$ vector, encapsulating $\mathbf{p}$ where $n_l$ is the total number of lattice nodes.
The lattice $\mathbf{s}$ contains $\mathbf{p}$. Specifically, the metric length of the lattice $\mathbf{s}$ in each spatial dimension is chosen slightly larger than the length of the object $\mathbf{p}$. 
As the lattice will represent the deformation of the object, its orientation should be set in a way that the lattice axes be aligned with the main axes that the object bends around. 
Furthermore, the resolution of $\mathbf{s}$ is selected high enough in each direction to be able to represent the deformation of the object in that direction effectively.
Typically $n_l$ can be chosen much smaller than $n$, i.e., the lattice is a compact representation of the object that allows for faster processing.
We generate a tetrahedral mesh over the lattice nodes in $\mathbf{s}$.
 An example of the formed lattice and the tetrahedral mesh around a shoe sole is illustrated in figure \ref{fig:lattice}. 
We form the set $\mathcal{T}_j$ for each point $\mathbf{p}_j$ in $\mathbf{p}$ as the indices of lattice nodes belonging to the tetrahedral cell encapsulating $\mathbf{p}_j$. 
We can, then, express $\mathbf{p}_j$ as a linear combination of lattice nodes in $\mathcal{T}_j$ using barycentric coordinates $\alpha_{i,j}$ as follows:
\begin{equation}
\label{eq:barycentric}
\sum_{\forall i\in \mathcal{T}_j}\alpha_{i,j} \mathbf{s}_{i} = \mathbf{p}_{j}.
\end{equation}
This equation serves as a geometrical constraint between $\mathbf{p}$ and $\mathbf{s}$. 
In the rest of the paper, by the shape of object and lattice, we refer to the 3D coordinates of the points in $\mathbf{p}$ and $\mathbf{s}$, respectively.
We also define the following superscripts that will henceforth be used to specify the state of different shapes:
\begin{itemize}
\item Current shape $c$
\item Desired shape $*$
\item Undeformed shape $u$
\end{itemize}
In the following sections, we describe the proposed tracking and servoing pipelines. 

\section{Tracking pipeline} \label{sec:tracking_pipeline}
\begin{figure}[t]\centering
    \includegraphics[trim={0 0 0 0},width=\linewidth]{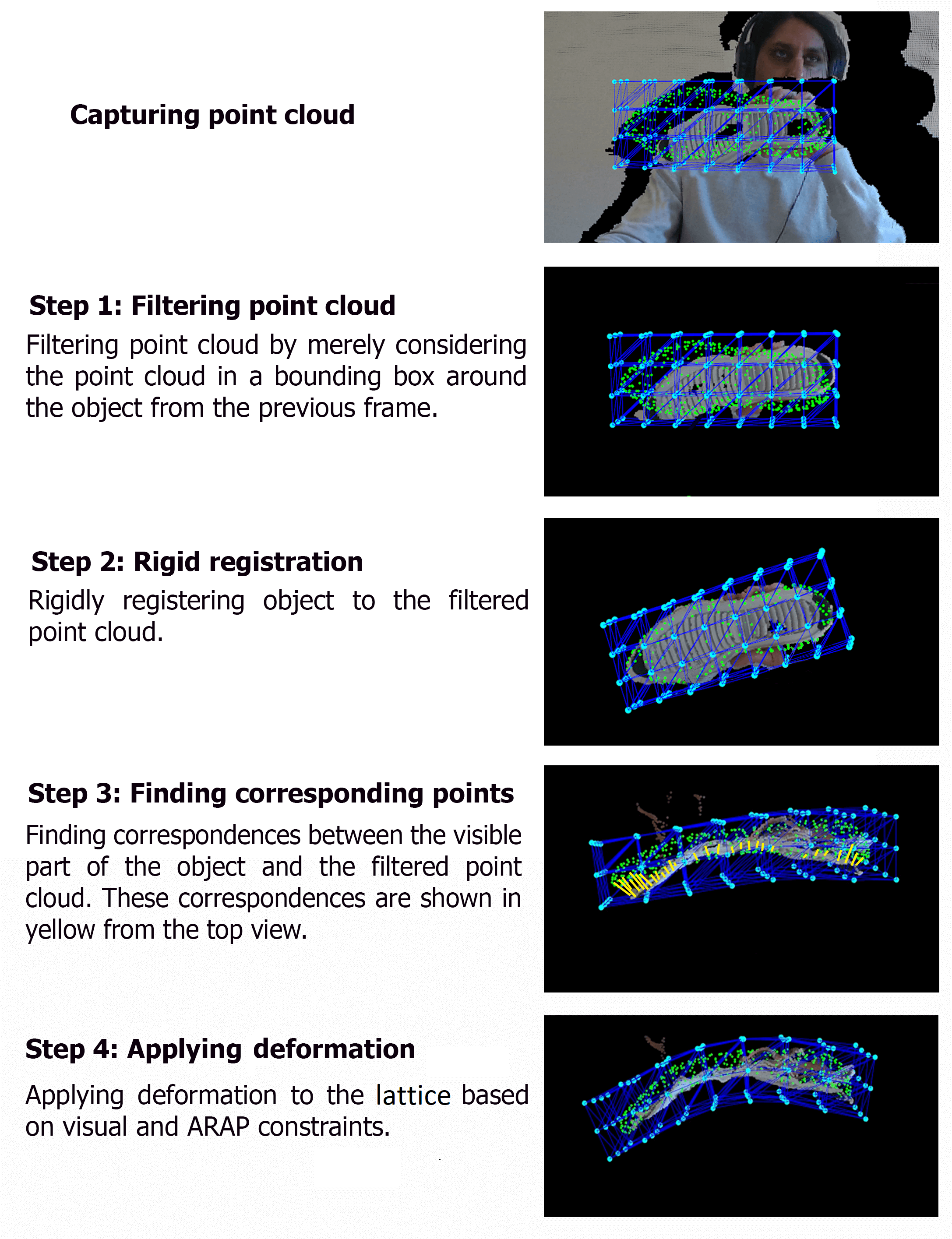} 
    \caption{ The results of different steps of the tracking pipeline in one frame of tracking a shoe sole. $\mathbf{p}^{c}$ and $\mathbf{s}^{c}$ are shown as green and light blue dots. The dark blue lines represent the tetrahedral mesh over $\mathbf{s}^{c}$.}
    \label{fig:tracking_results}
\end{figure}

Our proposed tracking pipeline is able to track a 3D elastic  deformable object in real time. 
It comprises 4 steps, namely: capturing and filtering point cloud, rigid registration, finding corresponding points, and finally, applying the deformation model. The first three steps of the pipeline are inspired by \cite{petit2015real}. The results of different steps of the tracking pipeline in tracking a shoe sole are shown in figure \ref{fig:tracking_results}.

The pipeline tracks the shape of the object in each frame from a known initial shape in 3D space. 
Knowing the initial shape of the object at the beginning of tracking is common in the state-of-the-art \cite{collins2015poster, petit2015real, ngo2015template, tang2018track}. We initially set the object and lattice shapes ($\mathbf{p}^{c}$ and $\mathbf{s}^{c}$) by rigidly transforming $\mathbf{p}^{u}$ (and thus $\mathbf{s}^{u}$) to an initial pose (i.e., a reference shape) in 3D space visible from the camera. 
For starting the tracking pipeline, we approximately align the object with the reference shape and trigger tracking. The alignment can be done by hand or by robots while grasping the object.
To facilitate the process of alignment, we visualize the reference shape in 3D space and its projection in 2D.
It should be noted that this alignment only needs to be partially consistent. This means that even if the real object is slightly displaced and deformed with respect to the reference shape, the tracking pipeline is able to infer a correct $\mathbf{p}^{c}$ after several frames. This is mainly thanks to the second step of our pipeline, i.e., rigid registration. We will explain this further in Sect. \ref{sec:rigid_registration}. 

\subsection{Step 1: Capturing and filtering point cloud} \label{sec:point_cloud_capturing_and_filtering}
The pipeline starts by capturing and filtering the point cloud from a 3D camera in each frame. 
The purpose of filtering is to remove the points in the point cloud that do not belong to the object but to the surroundings including background, other objects in the scene, and grippers.
The presence of these points in the point cloud might lead to disruption of the tracking pipeline, especially in step 3 (Sect. \ref{sec:finding_corresponding_points}) where correspondences will be found between the object and the captured point cloud.
We perform filtering by merely considering the point cloud inside a bounding box around $\mathbf{p}$ from the previous frame. In our experiments, we consider the dimensions of the bounding box to be $1$\,cm larger than the dimensions of $\mathbf{p}$ in each direction. We call this filtered point cloud $\mathbf{d}_{f}$. We also reduce the size of $\mathbf{d}_{f}$ by sampling the points on a $5$\,mm square grid. In addition to using a bounding box to filter the captured point cloud, one can use 2D image filters. This is optional but can be advantageous when there is a significant difference between the pixel characteristics of the object and its surroundings, e.g., a dark object in a light background.

\subsection{Step 2: Rigid registration } \label{sec:rigid_registration}
In this step, we rigidly register the points on the visible surface of $\mathbf{p}^{c}$, namely $\mathbf{p}_v^c$, to the points in $\mathbf{d}_{f}$. To this end, we exploit a classical rigid iterative closest point (ICP) algorithm. The output of this registration is rigid translation and rotation transformations that are applied on both $\mathbf{p}^{c}$ and $\mathbf{s}^{c}$. 
This step is essential to deal with rapid movements of the objects as it compensates for the rigid non-alignment between $\mathbf{p}^{c}$ and $\mathbf{d}_{f}$ in each frame. This is necessary to have a fair initialization for executing the next step of the tracking pipeline (i.e., finding corresponding points).
$\mathbf{p}_v^c$ is determined in each frame through a two-step process. First, the normals of $\mathbf{p}^{c}$ are calculated. Second, the points in $\mathbf{p}^{c}$ with an angle of more than 90 degrees between the normal of that point in $\mathbf{p}^{c}$ and the sightline passing through that point are selected. With this procedure, the selected points are in the visible surface of the object.
After rigidly registering the object, we update $\mathbf{p}_v^c$ to be used in the next step of the tracking pipeline.

\subsection{Step 3: Finding corresponding points} \label{sec:finding_corresponding_points}
In this step, we find correspondences between the points in $\mathbf{p}_v^c$ and $\mathbf{d}_{f}$. We use the ICP-like algorithm suggested in \cite{petit2015real}. In this algorithm, first, by employing K-d tree searches, nearest neighbor correspondences are determined, both from $\mathbf{p}_v^c$ to $\mathbf{d}_{f}$ and vice versa. Then, using these two sets of correspondences, a corresponding point in 3D space is computed for each point in $\mathbf{p}_v^c$. This is done by a two-step process: first, we average the coordinates of all the points in $\mathbf{d}_{f}$ having the same nearest neighbors in $\mathbf{p}_v^c$. 
Second, we average the resulting coordinates from the previous step and the coordinates of the points in $\mathbf{d}_{f}$ which are the nearest neighbors of each point of $\mathbf{p}_v^c$.
In this stage, we remove the points in $\mathbf{p}_v^c$ whose distance from their corresponding point is more than a specific threshold. 
We name the vector of 3D coordinates of these remaining points in $\mathbf{p}_v^c$ as $\mathbf{p}_a^c$ and the vector of their corresponding 3D coordinates in 3D space as $\mathbf{p'}_a^c$.
We indicate the number of points in $\mathbf{p}_a^c$ and $\mathbf{p'}_a^c$ with $n_a$.

\subsection{Step 4: Applying deformation} \label{sec:applying_deformation}
In this step, we apply deformation to the inferred shape. This deformation is applied to the lattice and not to the object and the visual constraints related to the object are imposed as the constraints to this deformation. This way, we solve for the shape of $\mathbf{s}^c$ and not $\mathbf{p}^c$, which is a faster process as $n_l$ is normally much smaller than $n$. It should be noted that $\mathbf{s}^c$ and $\mathbf{p}^c$ remain connected due to the spatial constraints in (\ref{eq:barycentric}).
We deform $\mathbf{s}^c$ by solving the following linear system, which is a modified version of the one suggested by \cite{zollhofer2012gpu}:

\begin{equation}
\label{eq:tracking}
\left ( \mathbf{\Gamma} \mathbf{L}+\mathbf{B}^\intercal\mathbf{B}\right )\mathbf{s}^c=\mathbf{\Gamma} \mathbf{b}+\mathbf{B}^\intercal\mathbf{t}.
\end{equation}
Two series of constraints are considered in this equation:

\begin{itemize}
\item \textit{ARAP constraints}. These constraints ensure that $\mathbf{s}^c$ and consequently $\mathbf{p}^c$ try to keep their local rigidity. These constraints are imposed by matrix $\mathbf{L}$ on the left-hand and vector $\mathbf{b}$ on the right-hand side of (\ref{eq:tracking}). $\mathbf{L}$ is the $n_l \times n_l$ Laplacian matrix of the tetrahedral mesh formed over the lattice, and $\mathbf{b}$ is a $n_l \times 3$ vector whose $i$th row is:
\begin{equation}
\label{eq:rhs_tracking}
\mathbf{b}_i = \sum_{j \in \mathcal{N}_i} \frac{w_{ij}}{2} (\mathbf{R}_i+\mathbf{R}_j)(\mathbf{s}^u_i-\mathbf{s}^u_j).
\end{equation}
In this equation, $\mathcal{N}_i$ is defined as the set containing the first-order neighbors of node $i$ in $\mathbf{s}$, $w_{ij}$ is a scalar encoding the connection between nodes $i$ and $j$ in $\mathbf{s}$, and $\mathbf{R}_i$ is the optimal rotation matrix that conforms the nodes of the set $\mathcal{N}_i$ in $\mathbf{s}^u$ to the same nodes in $\mathbf{s}^c$ with least-squares error. 
$\mathbf{R}_i$ is computed using the singular value decomposition (for more details see \cite{Sorkine2007}).

\item \textit{Object visual constraints}. These constraints try to minimize the error between the points in $\mathbf{p}_a^c$ and their corresponding points in $\mathbf{p'}_a^c$. They are imposed by $\mathbf{B}^\intercal\mathbf{B}$ on the left-hand and $\mathbf{B}^\intercal \mathbf{t}$ on the right-hand side of (\ref{eq:tracking}). $\mathbf{t}$ is a $n_a \times 3$ vector containing the 3D coordinates in $\mathbf{p'}_a^c$ as rows. $\mathbf{B}$ is an $n_a \times n_l$ matrix containing the constraints from (\ref{eq:barycentric}) as rows for each row of $\mathbf{t}$.
\end{itemize}
We adjust the effectiveness of these two constraints using $\mathbf{\Gamma}$. We consider $\mathbf{\Gamma}$ as a diagonal matrix with the size of $n_l \times n_l$ to attribute different values to each node of $\mathbf{s}^{c}$. This is concerned with the ARAP constraints as they strongly try to rigidly maintain the shape of $\mathbf{s}^{c}$ and consequently $\mathbf{p}^{c}$. We, thus, set two different values for each diagonal element of $\mathbf{\Gamma}$; 0.1 for the nodes of $\mathbf{s}^{c}$ belonging to a tetrahedral cell that surrounds a point of $\mathbf{p}_a^c$, and 1 for other nodes. As a result, the nodes in $\mathbf{s}^{c}$ being subject to deformation by object visual constraints can flexibly move, and thus the points in $\mathbf{p}_a^c$ can be driven toward their corresponding 3D coordinates in  $\mathbf{p'}_a^c$. At the same time, the other nodes in $\mathbf{s}^{c}$ keep the local rigidity, and thus the general shape of the lattice can be maintained.
Solving for $\mathbf{s}^{c}$ is not possible through a one-step solution. Instead as explained in \cite{zollhofer2012gpu}, an iterative flip-flop solution is used. 
In each flip-flop iteration, we calculate $\mathbf{R}_{i}$ (that is a function of $\mathbf{s}^{c}$) and use that in solving the linear system of (\ref{eq:tracking}) for $\mathbf{s}^{c}$.
We consider (\ref{eq:tracking}) converged when the difference between the two successive calculated $\mathbf{s}^{c}$ is smaller than a certain value.
In our experiments, the solution converges in 5-8 iterations. As $\mathbf{B}$ is sparse, we solve (\ref{eq:tracking}) for $\mathbf{s}^{c}$ using the conjugate gradient solver of Eigen library for sparse least-square problems.

Another constraint that can be incorporated into the tracking pipeline is the known 3D coordinates of several lattice nodes. These lattice nodes can be considered as rigidly attached to a known point in 3D space. In practice, this known point can be, e.g., a fixed point of the object or a robot’s gripper.
The indices of these lattice nodes are determined manually by the user before starting the tracking pipeline or automatically by selecting the closest lattice nodes to the known point in 3D space at the beginning of tracking.
In both cases, at the beginning of the tracking pipeline, the relative 3D coordinates of the selected lattice nodes with respect to their corresponding 3D known points are saved.
These relative coordinates along with the coordinates of the known points at each instant give us the 3D coordinates of the selected lattice nodes.
As for the grippers, their poses in the camera frame can be computed knowing the configuration of the robot and the calibration between the robots and the camera. 
The implementation of this constraint can be performed by modifying (\ref{eq:tracking}) before solving it. This is done by removing corresponding rows and columns from the left-hand side and recalculating the right-hand side with the known 3D coordinates of the relevant lattice nodes. The use of these constraints is optional; they make tracking more precise and robust, specifically in scenarios with large deformations or the presence of self or external occlusions. We provide illustrative videos of tracking-only experiments on the project website. 


\section{Servoing pipeline} \label{sec:servoing_pipeline}
In this section, we explain our shape servoing pipeline. The same as in our previous work \cite{shetab2022rigid}, we use the ARAP deformation model to obtain a deformation Jacobian. The control law we propose, then, is based on this Jacobian. In \cite{shetab2022rigid} we computed the Jacobian numerically, by simulating perturbations in every DOF of the grippers that grip the object. In contrast, here we use an analytical expression of the Jacobian. The advantages of this are: the Jacobian we compute does not involve any numerical approximation, and its computation is fully scalable as the number of grippers grows. Another main difference is that we derive the formulation for the lattice and not the object. This not only generalizes the servoing formulation for any form of the object, but also decouples the runtime complexity of the servoing from the objects’ geometric complexity.
We use this Jacobian to propose a control law. Finally, we apply the control law on the lattice to guide $\mathbf{s}^{c}$ toward $\mathbf{s}^{*}$ and consequently $\mathbf{p}^{c}$ toward $\mathbf{p}^{*}$ thanks to the spatial constraints between the lattice and the object in (\ref{eq:barycentric}). We present the different steps of the servoing pipeline in the following.

\subsection{ARAP deformation Jacobian}
The main ingredient of our shape servoing pipeline is the Jacobian that expresses how the infinitesimal gripper motions change the shape of the object. This is called \textit{deformation Jacobian} in the literature. 
Most shape servoing works \cite{berenson2013manipulation, Navarro2014, Navarro2018, Hu2018, Duenser2018, koessler2021efficient, Zhu2021,  shetab2022rigid, aghajanzadeh2022asap} assume that the object's shape changes quasi-statically and that this change can be represented by such deformation Jacobian. In this paper, we also make the same assumption. As in our previous work \cite{shetab2022rigid}, the main principle of our servoing pipeline is approximating the true deformation Jacobian of the object by the deformation Jacobian of the ARAP model of the object.
In particular, we present an analytic derivation of the deformation Jacobian of the ARAP model. 
We start with the main ARAP linear system
\begin{equation}
\label{eq:servoing_initial_equation}
\mathbf{L} \mathbf{s} = \mathbf{b}, 
\end{equation}
which is (\ref{eq:tracking}) with only deformation constraints. 
The principle of ARAP’s modeling \cite{Sorkine2007} is that the object’s shape (in our case the lattice's shape $\mathbf{s}$) in quasi-static equilibrium minimizes the ARAP deformation energy. In particular, this energy is at a local minimum under the existing constraints (e.g., regions of the object that are grasped and moved by a robot).
The gradient of this energy with respect to $\mathbf{s}$ has the following form:
\begin{equation}
\label{eq:servoing_gradient}
\mathbf{g} = \mathbf{L} \mathbf{s} - \mathbf{b}, 
\end{equation}
omitting multiplicative constants that are irrelevant to our purpose. $\mathbf{b}$ contains the $\mathbf{b}_i$ for each node $i$ in $\mathbf{s}$, which can be written as in (\ref{eq:rhs_tracking}). 
As the stable shape $\mathbf{s}$ is a local minimum, one makes the gradient zero to find $\mathbf{s}$: this is what (\ref{eq:servoing_initial_equation}) expresses.
In the standard ARAP formulation, the optimal rotations $\mathbf{R}_i$ in (\ref{eq:rhs_tracking}) depend on $\mathbf{s}$ via SVD. Note that in this section $\mathbf{s}$ represents any shape in the neighborhood of $\mathbf{s}^c$. This is because what we want to do is to compute the deformation Jacobian at $\mathbf{s}^c$. Then, our insight is that the optimal rotations for $\mathbf{s}$ will also be in the neighborhood of the optimal rotations for $\mathbf{s}^c$. Therefore, we can express the relative rotation between them as an infinitesimal rotation. We will show that this infinitesimal rotation can be parameterized as a linear function of the node positions, avoiding the SVD. With this linear parameterization, we will transform (\ref{eq:servoing_gradient}) in an expression where the dependency on $\mathbf{s}$ is fully linear. From this expression we will derive the deformation Jacobian. 

We start by writing the rotation matrix $\mathbf{R}_i$ in (\ref{eq:rhs_tracking}) as the multiplication of a general rotation matrix $\mathbf{R}^{o}_i$ and an infinitesimal rotation matrix $\mathbf{R}^{s}_i$:
\begin{equation}
\label{eq:sorkine_rotation_devision}
\mathbf{R}_i = \mathbf{R}^{s}_i \mathbf{R}^{o}_i.
\end{equation}
$\mathbf{R}^{o}_i$ is the optimal rotation matrix that best conforms the nodes of the set $\mathcal{N}_i$ in $\mathbf{s}^u$ to the same nodes in $\mathbf{s}^c$. 
Note that this rotation matrix $\mathbf{R}^{o}_i$ is known for us: it is the last optimal rotation matrix computed by our tracking pipeline.
We are computing the Jacobian for changes of shape in the neighborhood of the current shape. Therefore, for the Jacobian computation,  $\mathbf{R}^{o}_i$ is considered a fixed matrix.
$\mathbf{R}^{s}_i$, the infinitesimal rotation matrix, can be expressed as a linear function of $\mathbf{s}_i$ using the estimation suggested by \cite{sorkine2004laplacian}.
According to \cite{sorkine2004laplacian}, when $\mathbf{s}_i$ is in the local neighborhood of $\mathbf{s}^c_i$, one can write an approximation for the transformation matrix $\mathbf{T}_i$ between $\mathbf{s}^c_i$ and $\mathbf{s}_i$ that is a linear function of $\mathbf{s}_i$.
The transformation matrix $\mathbf{T}_i$ is a $4\times 4$ matrix including translation, rotation, and scale. Here, what we need is the rotation part of this transformation. We, consequently, isolate the rotation part of the $\mathbf{T}_i$ formulation presented in \cite{sorkine2004laplacian} and use it as $\mathbf{R}^{s}_i$.
We can thus write $\mathbf{R}^{s}_i$ as the following infinitesimal rotation matrix:
\begin{equation}
\label{eq:rotation_sorkine}
\mathbf{R}^{s}_i = 
\begin{pmatrix}
1 & -h_{i_3} & h_{i_2}\\ 
h_{i_3} & 1 & -h_{i_1}\\ 
-h_{i_2} & h_{i_1} & 1
\end{pmatrix}.
\end{equation}
As this is a least-squares optimal rotation, $h_{i_1}$, $h_{i_2}$, and $h_{i_3}$ can be computed from:
\begin{equation}
\label{eq:sorkine_h_calculation}
\begin{pmatrix}
h_{i_1}\\ 
h_{i_2}\\ 
h_{i_3}
\end{pmatrix}
=
(\mathbf{A}^\intercal_i \mathbf{A}_i)^{-1} \mathbf{A}^\intercal_i \mathbf{w}_i.
\end{equation}
We consider that the reference shape for this least-squares computation is $\mathbf{s}^{c}$. Therefore, $\mathbf{A}_i$ and $\mathbf{w}_i$ can be written as:
\begin{equation}
\label{eq:sorkine_A_calculation}
\mathbf{A}_i =
\begin{pmatrix}
0 & \mathbf{s}^{c}_{k_z} & -\mathbf{s}^{c}_{k_y}\\ 
-\mathbf{s}^{c}_{k_z} & 0 & \mathbf{s}^{c}_{k_x}\\ 
\mathbf{s}^{c}_{k_y} & -\mathbf{s}^{c}_{k_x} & 0 \\ 
\vdots  & \vdots & \vdots
\end{pmatrix},k\in \begin{Bmatrix}
i
\end{Bmatrix}\cup \mathcal{N}_i,
\end{equation}
and 
\begin{equation}
\label{eq:sorkine_b_calculation}
\mathbf{w}_i =
\begin{pmatrix}
\mathbf{s}_{k_x} \\ 
\mathbf{s}_{k_y} \\ 
\mathbf{s}_{k_z} \\ 
\vdots
\end{pmatrix}
,k\in \begin{Bmatrix}
i
\end{Bmatrix}\cup \mathcal{N}_i.
\end{equation}
$\mathbf{A}_i$ is a known $3d_i\times 3$ matrix ($d_i$ is the number of elements in $\begin{Bmatrix}i\end{Bmatrix}\cup \mathcal{N}_i$) composed of the elements from the current shape, and $\mathbf{w}_i$ is a $3d_i\times 1$ vector of unknowns comprising $\mathbf{s}_i$ and its first-order neighbors. By putting (\ref{eq:sorkine_A_calculation}) and (\ref{eq:sorkine_b_calculation}) in (\ref{eq:sorkine_h_calculation}) and then in (\ref{eq:rotation_sorkine}) we can have $\mathbf{R}^s_i$ as: 
\begin{equation}
\label{eq:rotation_sorkine_replaced}
\mathbf{R}^{s}_i = 
\begin{pmatrix}
1 & -\mathbf{M}_{i_{3,*}}\mathbf{w}_i & \mathbf{M}_{i_{2,*}}\mathbf{w}_i\\ 
\mathbf{M}_{i_{3,*}}\mathbf{w}_i & 1 & -\mathbf{M}_{i_{1,*}}\mathbf{w}_i\\ 
-\mathbf{M}_{i_{2,*}}\mathbf{w}_i & \mathbf{M}_{i_{1,*}}\mathbf{w}_i & 1
\end{pmatrix},
\end{equation}
where $\mathbf{M}_i = (\mathbf{A}^\intercal_i \mathbf{A}_i)^{-1} \mathbf{A}^\intercal_i$ is a $3\times3d_i$ matrix and the subscript $k,*$ indicates the $k$th row of the matrix $\mathbf{M}_i$.
We rearrange $\mathbf{R}^{s}_i$ as the following: 
\begin{equation}
\label{eq:rotation_sorkine_replaced2}
\mathbf{R}^{s}_i = 
\mathbf{I} + 
\begin{pmatrix}
\mathbf{0}_{3d_i}\mathbf{w}_i & -\mathbf{M}_{i_{3,*}}\mathbf{w}_i & \mathbf{M}_{i_{2,*}}\mathbf{w}_i\\ 
\mathbf{M}_{i_{3,*}}\mathbf{w}_i & \mathbf{0}_{3d_i}\mathbf{w}_i & -\mathbf{M}_{i_{1,*}}\mathbf{w}_i\\ 
-\mathbf{M}_{i_{2,*}}\mathbf{w}_i & \mathbf{M}_{i_{1,*}}\mathbf{w}_i & \mathbf{0}_{3d_i}\mathbf{w}_i
\end{pmatrix},
\end{equation}
where $\mathbf{I}$ is a $3\times 3$ identity matrix, and $\mathbf{0}_{3d_i}$ is a row vector of zeros with the length of $3d_i$.
Replacing (\ref{eq:rotation_sorkine_replaced2}) in (\ref{eq:sorkine_rotation_devision}) and then in (\ref{eq:rhs_tracking}) and rearranging the terms we have:
\begin{equation}
\label{eq:rhs_larap_servoing_new}
\mathbf{b}_i = \sum_{j \in \mathcal{N}_i} \frac{w_{ij}}{2}\mathbf{I} (\mathbf{R}^o_i+\mathbf{R}^o_j)(\mathbf{s}^u_i-\mathbf{s}^u_j) +
\sum_{j \in \mathcal{N}_i} \mathbf{q}_{i} \mathbf{w}_i + 
\sum_{j \in \mathcal{N}_i} \mathbf{q}_{j} \mathbf{w}_j,
\end{equation}
where $\mathbf{q}_{i}$ and $\mathbf{q}_{j}$ are known $1\times 3d_i$ and $1\times 3d_j$ vectors that can be written as the following: 
\begin{equation}
\begin{split}
\label{eq:rhs_larap_servoing_qij}
\mathbf{q}_{i} = \frac{w_{ij}}{2} (
\mathbf{u}_i[1](\mathbf{M}_{i_{3,*}}-\mathbf{M}_{i_{2,*}})+ \\
\mathbf{u}_i[2](\mathbf{M}_{i_{1,*}}-\mathbf{M}_{i_{3,*}})+ \\
\mathbf{u}_i[3](\mathbf{M}_{i_{2,*}}-\mathbf{M}_{i_{1,*}}) ),
\end{split}
\end{equation}
\begin{equation}
\begin{split}
\label{eq:rhs_larap_servoing_qjj}
\mathbf{q}_{j} = \frac{w_{ij}}{2} (
\mathbf{u}_j[1](\mathbf{M}_{j_{3,*}}-\mathbf{M}_{j_{2,*}})+ \\
\mathbf{u}_j[2](\mathbf{M}_{j_{1,*}}-\mathbf{M}_{j_{3,*}})+ \\
\mathbf{u}_j[3](\mathbf{M}_{j_{2,*}}-\mathbf{M}_{j_{1,*}}) ).
\end{split}
\end{equation}
In these equations $\mathbf{u}_i = \mathbf{R}^o_i(\mathbf{s}^u_i-\mathbf{s}^u_j)$ and $\mathbf{u}_j = \mathbf{R}^o_j(\mathbf{s}^u_i-\mathbf{s}^u_j)$ are both $3\times1$ vectors, and $[k]$ signifies the $k$th element of the vector.

On the right-hand side of (\ref{eq:rhs_larap_servoing_new}), the first sum is a known and constant vector, while the second and third sums are linear functions of $\mathbf{w}_i$ and $\mathbf{w}_j$.
One point that should be noted is that for solving the original formulation of ARAP in (\ref{eq:servoing_initial_equation}), as described in Sect. \ref{sec:applying_deformation}, a flip-flop solution is used. This solution comprises two steps: calculating optimal rotations and solving the ARAP linear system. In calculating optimal rotations, the directions of $x$, $y$, and $z$ are dependent in the SVD. In solving the ARAP linear system, however, having the optimal rotation matrices calculated, one can solve (\ref{eq:servoing_initial_equation}) for each direction independently regardless of the other directions.
In our new formulation of $\mathbf{b}$ in (\ref{eq:rhs_larap_servoing_new}), we propagate the dependency between these directions (coming from the second and third sums) into the equation.
This dependency stems from the usage of the estimation for rotation matrices (see equations (\ref{eq:rotation_sorkine}) to (\ref{eq:sorkine_b_calculation})). 
In order to integrate this dependency with other terms in the original formulation of ARAP, we rewrite each term in (\ref{eq:servoing_initial_equation}).
This is done as follows:

\begin{itemize}
\item 
\textit{(i)} We define $\mathbf{s}'$ as the flattened form of $\mathbf{s}$ with the size of $3n_l \times 1$.
\item 
\textit{(ii)} We define $\mathbf{L}'$ as a $3n_l \times 3n_l$ matrix such that $\mathbf{L}'=\mathbf{L} \otimes \mathbf{I}$ where $\mathbf{I}$ is a $3\times 3$ identity matrix and $\otimes$ denotes the Kronecker product.
\item 
\textit{(iii)} We define $\mathbf{b}'$ as the following:
\begin{equation}
\label{eq:sorkine_b}
    \mathbf{b}' = \mathbf{c} + \mathbf{Q} \mathbf{s}'.
\end{equation}
In this equation, $\mathbf{b}'$ with the size of $3n_l \times 1$ is the flattened form of $\mathbf{b}$ with the size of $n_l \times 3$.
$\mathbf{c}$ is a $3n_l \times 1$ vector that represents the general form of the first sum of (\ref{eq:rhs_larap_servoing_new}) consisting of $n_l$ vectors of all lattice nodes (each of size $3\times1$) concatenated together.
The multiplication of $\mathbf{Q} \mathbf{s}'$ includes all the elements in the second and third sums in (\ref{eq:rhs_larap_servoing_new}).
The same as in \textit{(i)}, $\mathbf{s}'$ is the flattened form of $\mathbf{s}$ with the size of $3n_l \times 1$.
$\mathbf{Q}$ is a $3n_l \times 3n_l$ known matrix that is initially filled with zeros and then the values from $\mathbf{q}_{i}$ and $\mathbf{q}_{j}$ will be added to their corresponding elements.
Each consecutive three rows in $\mathbf{Q}$ belong to three directions of one lattice node. We set these three rows identical. This comes from the fact that we apply the constraints from the second and the third sums in (\ref{eq:rhs_larap_servoing_new}) that depend on the three directions to each one of the directions identically.
Each consecutive three columns of $\mathbf{Q}$ are dedicated to the three directions of each lattice node. 
The elements in $\mathbf{q}_{i}$ and $\mathbf{q}_{j}$ are summed with the elements in the rows corresponding to the $i$th lattice node at their corresponding columns regarding their indices and their directions.
The process of filling $\mathbf{Q}$ can be formulated by the following equation:
\begin{equation}
\begin{split}
\label{eq:Q_calculation}
    \mathbf{Q}(3(i-1)+r,3(\mathcal{K}[k]-1)+c) = \sum_{j\in \mathcal{N}_i}\mathbf{q}[3(k-1)+c], 
\end{split}
\end{equation}
where $\mathbf{q}\in \left \{ \mathbf{q}_i, \mathbf{q}_j \right \}$, $i\in \left \{ 1,...,n_l \right \}$, $r,c\in \left \{ 1,2,3 \right \}$ and

\begin{equation}
\begin{split}
\label{eq:Q_calculation2}
    \left\{\begin{matrix}
        k\in \left \{ 1,...,d_i \right \}, \mathcal{K} = i\cup \mathcal{N}_i, &   \text{if} \; \mathbf{q} = \mathbf{q}_i
    \\ 
        k\in \left \{ 1,...,d_j \right \}, \mathcal{K} = j\cup \mathcal{N}_j,  &  \text{if} \; \mathbf{q} = \mathbf{q}_j
\end{matrix}\right.
\end{split}.
\end{equation}

\end{itemize}
Using flattening and (\ref{eq:sorkine_b}), we can write (\ref{eq:servoing_gradient}) as:
\begin{equation}
\label{eq:sorkine_final_servoing_equation3}
    \mathbf{g}'= \mathbf{L}' \mathbf{s}' - \mathbf{b}'= \mathbf{H} \mathbf{s}' - \mathbf{c},
\end{equation}
where $\mathbf{g}'$ with the size of $3n_l$ is the flattened form of $\mathbf{g}$, and $\mathbf{H}=\mathbf{L}'-\mathbf{Q}$ is a known $3n_l \times 3n_l$ matrix.
We will define the deformation Jacobian in terms of velocities (i.e., time variations). 
Note that for this Jacobian computation, $\mathbf{H}$ and $\mathbf{c}$ are constant, as they depend on fixed quantities. 
Therefore, by taking derivative from (\ref{eq:sorkine_final_servoing_equation3}) with respect to time, we have $\mathbf{H} \dot{\mathbf{s}}' = \dot{\mathbf{g}}'$.

The final step of our derivation is to obtain the deformation Jacobian from this equation. For this, we apply on our ARAP model an approach that is based on node partitioning. This approach has been previously applied in \cite{koessler2021efficient} on an FEM model, and in \cite{aghajanzadeh2022asap} on an offline geometrical model.
In \cite{koessler2021efficient, aghajanzadeh2022asap} this approach was defined for linear objects deforming in 2D; here, we widen that scope as we consider objects of arbitrary form (linear, thin-shell, volumetric) which deform in 3D.

We, first, categorize the nodes of the lattice into three sets; free, servoed, and gripped, having in turn $n_f$, $n_s$, and $n_g$ elements such that $n_f + n_s + n_g = n_l$.
Accordingly, we divide the flattened position vector of the lattice nodes $\mathbf{s}'$ into $\mathbf{s}'_f\in \mathbb{R}^{3n_f}$, $\mathbf{s}'_s\in \mathbb{R}^{3n_s}$, and $\mathbf{s}'_g\in \mathbb{R}^{3n_g}$. Likewise, we partition matrix $\mathbf{H}$. 
Under ARAP modeling, the gripped nodes are moved externally, but the motions of the free and servoed nodes are determined only by the ARAP deformation energy. As the object is always in quasi-static equilibrium, its shape corresponds to a locally minimum energy. Therefore, for the free and servoed nodes the ARAP energy gradient has to be always zero. Hence, the time derivative of the gradient is also zero. Therefore, the expression $\mathbf{H} \dot{\mathbf{s}}' = \dot{\mathbf{g}}'$ above takes the following form:

\begin{equation}
\label{eq:servoing_partitioned}
    \begin{pmatrix}
\mathbf{H}_{gg} & \mathbf{H}_{gs} & \mathbf{H}_{gf}\\ 
\mathbf{H}_{sg} & \mathbf{H}_{ss} & \mathbf{H}_{sf}\\ 
\mathbf{H}_{fg} & \mathbf{H}_{fs} & \mathbf{H}_{ff} 
\end{pmatrix}\begin{pmatrix}
\dot{\mathbf{s}}'_g\\ 
\dot{\mathbf{s}}'_s\\ 
\dot{\mathbf{s}}'_f
\end{pmatrix}
= 
\begin{pmatrix}
\dot{\mathbf{g}}'_g\\ 
\mathbf{0}\\ 
\mathbf{0}
\end{pmatrix}.
\end{equation}
From (\ref{eq:servoing_partitioned}), we can obtain the following expression linking the velocities of the gripped and the servoed nodes:
\begin{equation}
\label{eq:gripped_servoed_relation}
    \dot{\mathbf{s}}'_s = \mathbf{J}_{sg}\dot{\mathbf{s}}'_g, 
\end{equation}
where 
\begin{equation}
\label{eq:J_sg}
    \mathbf{J}_{sg}=-(\mathbf{H}_{ss}-\mathbf{H}_{sf}\mathbf{H}_{ff}^{-1}\mathbf{H}_{fs})^{-1}(\mathbf{H}_{sg}-\mathbf{H}_{sf}\mathbf{H}_{ff}^{-1}\mathbf{H}_{fg}). 
\end{equation}
As in \cite{koessler2021efficient, yu2022shape, aghajanzadeh2022asap}, we assume the matrices that have to be inverted are full-rank. This stems from the initial made assumption that the shape is fully constrained by the grippers.
Thus, $\mathbf{J}_{sg}$ is the ARAP deformation Jacobian, which we have obtained analytically from the knowledge of the current shape $\mathbf{s}^c$, the optimal rotations $\mathbf{R}^{o}$, and the ARAP model parameters.

\subsection{Control law for shape servoing}
We use $\mathbf{J}_{sg}$ to propose a proportional control law to drive the positions of the servoed nodes of the lattice toward their desired values.
We, first, define the servoing error as:
\begin{equation}
\label{eq:servoing_error}
    \mathbf{e}_s = \mathbf{s}'^c_s - \mathbf{s}'^*_s.
\end{equation}
Using (\ref{eq:gripped_servoed_relation}) and (\ref{eq:servoing_error}) we can write our proportional control law as the following:
\begin{equation}
\label{eq:control_law}
    \mathbf{v}_g = -k_{g}\mathbf{J}^\dagger_{sg}\mathbf{e}_s,
\end{equation}
where $k_{g}$ is a positive gain, $\dagger$ signifies the pseudoinverse, and $\mathbf{v}_g$ is the vector of translational velocities to be applied to the gripped lattice nodes.
In a common robotic application, the object is controlled by a robotic gripper with 6 DOFs. Hence, we transfer the calculated translational velocities of the gripped lattice nodes to the 6-DOF velocities of their corresponding grippers. Note that the grippers firmly hold the object and, thus, are coupled with the lattice due to the constraints in (\ref{eq:barycentric}).
We categorize the gripped nodes corresponding to each gripper in the sets $\mathcal{G}_l$ such that $\sum_{l=1}^{m}\left | \mathcal{G}_l \right |=n_g$ where $\left | . \right |$ signifies the number of nodes in the set.
We can, then, write the following equation between $\mathbf{v}_{p_l}$, the $6\times 1$ velocity vector of the gripper $l$, and $\mathbf{v}_{g_l}$, the velocity vectors of the nodes in $\mathcal{G}_l$ stacked together. 
\begin{equation}
\label{eq:gripped_h_servoed_relation}
    \mathbf{v}_{g_l} = \mathbf{G}_{gp_l}\mathbf{v}_{p_l},
\end{equation}
where $\mathbf{G}_{gp_l}$ is the grasp matrix of the gripper $l$ and can be written as:
\begin{equation}
\label{eq:jacobian_one_gripper}
\mathbf{G}_{gp_l} =
    \begin{pmatrix}
1 & 0 & 0 & 0 & {r_{lj}}_{z} & -{r_{lj}}_{y}\\ 
0 & 1 & 0 & -{r_{lj}}_{z} & 0 & {r_{lj}}_{x}\\
0 & 0 & 1 &  {r_{lj}}_{y} & -{r_{lj}}_{x} & 0\\ 
\vdots & \vdots & \vdots & \vdots & \vdots & \vdots
\end{pmatrix}
, j \in \mathcal{G}_l,
\end{equation}
where ${r_{lj}}_{x}$, ${r_{lj}}_{y}$, ${r_{lj}}_{z}$ represent the three components of the vector from the gripper $l$ to the lattice node $j$ at each instant. 
We extend (\ref{eq:jacobian_one_gripper}) to write a general equation for all the grippers and their corresponding lattice nodes:
\begin{equation}
\label{eq:gripped_patch_relation}
    \mathbf{v}_g = \mathbf{G}_{gp}\mathbf{v}_p,
\end{equation}
where $\mathbf{v}_g$ is a $3n_g\times 1$ vector containing the translational velocities of all the gripped nodes of the lattice stacked together, $\mathbf{v}_p$ is a $6m\times 1$ vector containing the velocities of all the grippers stacked together, and $\mathbf{G}_{gp}$ is the total grasp matrix with the size of $ 3n_g\times 6m$ assembled from all the grasp matrices $\mathbf{G}_{gp_l}$:
\begin{equation}
\label{eq:jacobian_gripper}
\mathbf{G}_{gp} =
    \begin{pmatrix}
\mathbf{G}_{gp_1} & 0 & \cdots & 0 \\ 
0 & \ddots &   &  \\
\vdots &   & \mathbf{G}_{gp_l} &  \\ 
0 &   &   & \ddots\\ 
\end{pmatrix}
, l \in \mathcal{M}.
\end{equation}
We finally define the control law:
\begin{equation}
\label{eq:control_law_final}
    \mathbf{v}_p = -k_{p}\mathbf{J}^\dagger_{sp}\mathbf{e}_s, 
\end{equation}
where $k_{p}$ is a positive gain, and $\mathbf{J}_{sp} = \mathbf{J}_{sg}\mathbf{G}_{gp}$ is the Jacobian that relates the lattice servoed nodes to the grippers.
One can use a diagonal gain matrix instead of $k_{p}$ to weight differently translation and rotation velocities.


In \cite{shetab2022rigid} we carried out shape servoing with a Jacobian-based proportional control law. In contrast to this previous work, our new control law in (\ref{eq:control_law_final}) is based on an analytic formulation of the deformation Jacobian, is applicable on objects of all forms, and can be used for both full and partial shape servoing. This new control law allows an exponential decrease of the shape servoing error $\mathbf{e}_s$ towards zero if the ARAP deformation Jacobian approximates the object's true deformation Jacobian well. The practical performance of the control law is illustrated and discussed in detail in the experiments section.

\section{Experiments}
In this section, we validate the effectiveness of our approach through a diverse set of experiments covering various forms and materials of elastic deformable objects.
The videos of these experiments and of experiments using only the tracking pipeline  can be found on the project website.
The objects of interest comprise a linear object, i.e., a cable, two thin-shell objects including a sheet of A4 paper and a convoluted foam, and two volumetric objects including a shoe sole and a foam octagonal cylinder. 
We apply large deformations and do both full and partial shape servoing.
Full shape servoing is when we servo all the lattice nodes (except for the gripped nodes), i.e., $n_f=0$.
Partial shape servoing is, however, when we servo a portion of the lattice nodes, i.e., $n_f\neq 0$.
The ability to perform partial shape servoing is interesting in practice (e.g., for tasks where a  specific part of the object has to be assembled on another object), and it highlights the versatility of our approach.

\subsection{Experimental setup} \label{subsec:experimental_setup}
The experiments are conducted using a dual-arm setup made of two Franka Emika Panda robots each with 7 DOFs. A RealSense D435 camera facing the manipulation area provides the input for the tracking pipeline. We use the camera resolution $424\times240$ in all the experiments except for the ones with linear objects in which we double this resolution to apply image filters (see Sect. \ref{subsec:Linear_objects}). We use MoveIt Hand-eye calibration plugin to externally calibrate the camera with the robots. The setup is shown in figure \ref{fig:setup}.
The whole code is written in C++ and runs on a single ROS node on a Dell laptop with a 9th generation Intel Core i7 CPU. 
No parallelization is employed to run the ROS node.
Our approach calculates the grippers' velocities and sends them to a Cartesian velocity control ROS node that controls each robot.
We use Point Cloud Library (PCL) for handling point clouds. 

\begin{table*}[!t]
    \caption{Parameters of the shape servoing tasks.\label{tab:experiments}}
    \centering
    \resizebox{\textwidth}{!}{
    \begin{tabular}{cccccccc}
        Task & 
        Object Type & 
        Object & 
        \makecell{Template \\ creation} &
        \makecell{Object point \\ cloud size} & 
        \makecell{Lattice \\ Dimension} &
        \makecell{Full/partial \\ shape servoing} & 
        Main feature \\
        \hline
        \multicolumn{1}{c|}{T1.1} & 
        \multicolumn{1}{c|}{\multirow{3}{*}{Linear}}  & \multicolumn{1}{c|}{\multirow{3}{*}{Cable}} &
        \multicolumn{1}{c|}{\multirow{3}{*}{Blender}} &
        \multicolumn{1}{c|}{\multirow{3}{*}{1734}} &
        \multicolumn{1}{c|}{\multirow{3}{*}{15$\times$3$\times$3}} &
        f & 
        In-plane deformation, large deformation \\
        
        \multicolumn{1}{c|}{T1.2} & 
        \multicolumn{1}{c|}{} & 
        \multicolumn{1}{c|}{} & 
        \multicolumn{1}{c|}{} & 
        \multicolumn{1}{c|}{} & 
        \multicolumn{1}{c|}{} & 
        p & 
        In-plane deformation \\
        
        \multicolumn{1}{c|}{T1.3} & 
        \multicolumn{1}{c|}{} & 
        \multicolumn{1}{c|}{} & 
        \multicolumn{1}{c|}{} & 
        \multicolumn{1}{c|}{} & 
        \multicolumn{1}{c|}{} & 
        f & 
        Out-of-plane deformation  \\
        
        \hline
        
        \multicolumn{1}{c|}{T2.1} & 
        \multicolumn{1}{c|}{\multirow{6}{*}{Thin-shell}}  & \multicolumn{1}{c|}{\multirow{4}{*}{A4 Paper}} & 
        \multicolumn{1}{c|}{\multirow{4}{*}{Blender}} &
        \multicolumn{1}{c|}{\multirow{4}{*}{1024}} &
        \multicolumn{1}{c|}{\multirow{4}{*}{8$\times$8$\times$3}} &
        f & 
        Large deformation \\
        
        \multicolumn{1}{c|}{T2.2} & 
        \multicolumn{1}{c|}{} & 
        \multicolumn{1}{c|}{} & 
        \multicolumn{1}{c|}{} & 
        \multicolumn{1}{c|}{} & 
        \multicolumn{1}{c|}{} & 
        p & 
        Separated servoing regions  \\
        
        \multicolumn{1}{c|}{T2.3} & 
        \multicolumn{1}{c|}{} & 
        \multicolumn{1}{c|}{} & 
        \multicolumn{1}{c|}{} & 
        \multicolumn{1}{c|}{} & 
        \multicolumn{1}{c|}{} & 
        p & 
        Only translation  \\

        \multicolumn{1}{c|}{T2.4} & 
        \multicolumn{1}{c|}{} & 
        \multicolumn{1}{c|}{} & 
        \multicolumn{1}{c|}{} & 
        \multicolumn{1}{c|}{} & 
        \multicolumn{1}{c|}{} & 
        p & 
        Changing grippers' positions on the object  \\
        
        \cline{3-8}
        
        \multicolumn{1}{c|}{T2.5} & 
        \multicolumn{1}{c|}{} & 
        \multicolumn{1}{c|}{\multirow{2}{*}{Convoluted foam}} & 
        \multicolumn{1}{c|}{\multirow{2}{*}{Scan}} & 
        \multicolumn{1}{c|}{\multirow{2}{*}{5535}} & 
        \multicolumn{1}{c|}{\multirow{2}{*}{8$\times$8$\times$3}} & 
        f & 
        Only translation, large deformation  \\
        
        \multicolumn{1}{c|}{T2.6} & 
        \multicolumn{1}{c|}{} & 
        \multicolumn{1}{c|}{} & 
        \multicolumn{1}{c|}{} & 
        \multicolumn{1}{c|}{} & 
        \multicolumn{1}{c|}{} & 
        p & 
        Only translation, small separated servoing regions  \\
        
        \hline
        
        \multicolumn{1}{c|}{T3.1} & 
        \multicolumn{1}{c|}{\multirow{4}{*}{Volumetric}}  & \multicolumn{1}{c|}{\multirow{2}{*}{Shoe sole}} & 
        \multicolumn{1}{c|}{\multirow{2}{*}{Scan}} & 
        \multicolumn{1}{c|}{\multirow{2}{*}{502}} & 
        \multicolumn{1}{c|}{\multirow{2}{*}{8$\times$4$\times$4}} & 
        f & 
        Large deformation \\
        
        \multicolumn{1}{c|}{T3.2} & 
        \multicolumn{1}{c|}{} & 
        \multicolumn{1}{c|}{} & 
        \multicolumn{1}{c|}{} & 
        \multicolumn{1}{c|}{} & 
        \multicolumn{1}{c|}{} & 
        f & 
        Severe rotation, change of view  \\
        
        \cline{3-8}
        
        \multicolumn{1}{c|}{T3.3} & 
        \multicolumn{1}{c|}{} & 
        \multicolumn{1}{c|}{\multirow{2}{*}{Foam octagonal cylinder}} & 
        \multicolumn{1}{c|}{\multirow{2}{*}{Blender}} & 
        \multicolumn{1}{c|}{\multirow{2}{*}{2352}} & 
        \multicolumn{1}{c|}{\multirow{2}{*}{8$\times$4$\times$4}} & 
        f & 
        Twist  \\
        
        \multicolumn{1}{c|}{T3.4} & 
        \multicolumn{1}{c|}{} & 
        \multicolumn{1}{c|}{} & 
        \multicolumn{1}{c|}{} & 
        \multicolumn{1}{c|}{} & 
        \multicolumn{1}{c|}{} & 
        f & 
        Twist + Bending deformation  \\
        
        \hline
    \end{tabular}}
\end{table*}

\begin{figure*}[!t]\centering
    \includegraphics[trim={0 10pt 0 10pt},width=.99\linewidth]{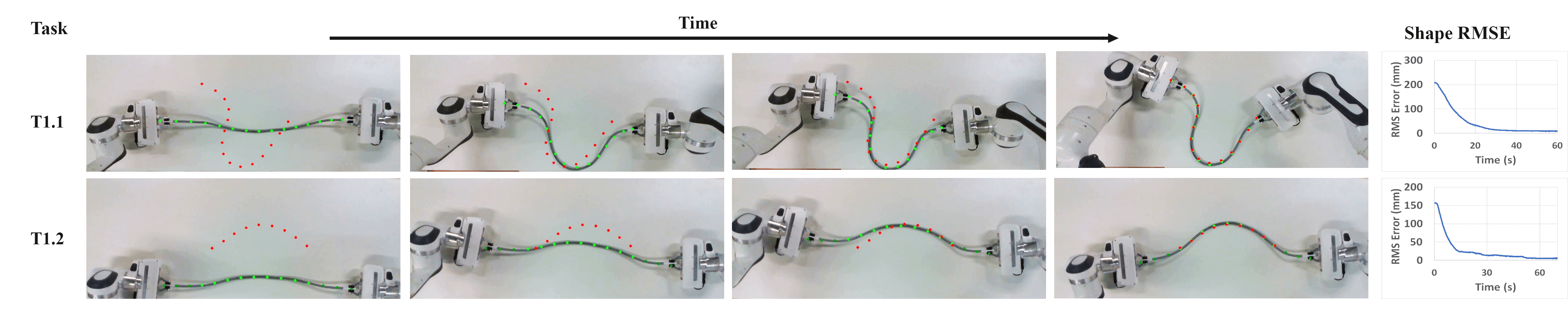} 
    \caption{Tasks with a cable and in-plane deformations. The plotted points represent the mean coordinates of the lattice nodes belonging to the cross-sections of the lattice along the cable length. Green points: current lattice shape, red points: desired lattice shape. Top row: full shape servoing, bottom row: partial shape servoing.}
    \label{fig:experiments_linear}
\end{figure*}

\begin{figure*}[!t]\centering
    \includegraphics[trim={0 10pt 0 10pt},width=.99\linewidth]{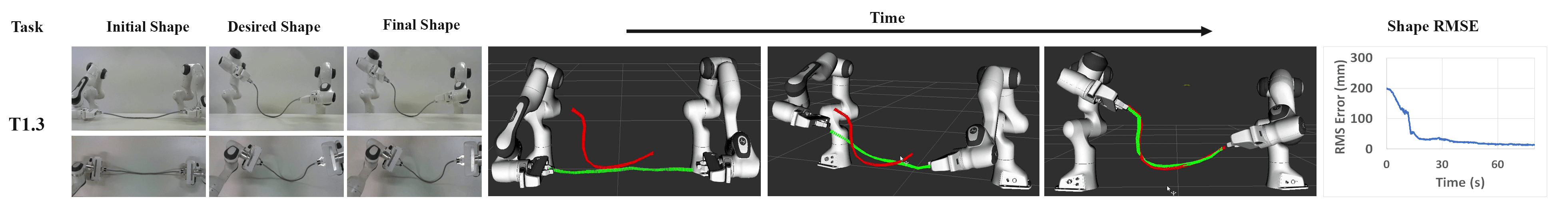} 
    \caption{Task with a cable and out-of-plane deformations. Green: cable's current shape, red: cable's desired shape.}
    \label{fig:experiments_linear_outofplane}
\end{figure*}

Depending on the complexity of the objects' geometry, we form the undeformed point cloud of each object $\mathbf{p}^u$ either by scanning them (using a Kinect and Skanect software) or by drawing simple shapes in Blender.
We, then, form the lattice $\mathbf{s}^u$ around each object point cloud.
In all cases, we consider the size of the lattice to be 1\,cm larger than the size of the object in each direction. 
We rigidly transform the created $\mathbf{p}^u$, as the initial shape $\mathbf{p}^c$, somewhere in front of the camera where it is reachable by the robots. We, then, align the real object with the initial shape $\mathbf{p}^c$ while being grasped by the robots.
Note that a partial alignment is sufficient (see Sect. \ref{sec:tracking_pipeline}). 

We, then, trigger the tracking pipeline. 
After the tracking pipeline successfully starts tracking the shape of the object, we activate the use of grippers' 3D coordinates in the tracking pipeline. 
We select the eight closest lattice nodes to each gripper as the lattice nodes with known coordinates to be used in the tracking pipeline.
In this stage, we also select the gripped lattice nodes in the servoing pipeline. 
We use the same selected lattice nodes with the known coordinates in the tracking pipeline as the gripped lattice nodes.
Next, we set the lattice desired shape $\mathbf{s}^*$. To this end, we manually deform the object by moving the robotic arms while grasping the object.
This is a natural way of defining the desired shape. One can also define the desired shape without the robots holding the object; our approach does not have any constraint in this respect.
We store the lattice desired shape $\mathbf{s}^*$ which is corresponding to the object's desired shape $\mathbf{p}^*$.
The next step is to manually move the robots to set the initial shapes of the lattice and the object. This is done in the same way as setting the desired shapes.
Finally, we start the servoing pipeline to drive the lattice (and thus the object) from its initial shape to the desired shape.
As a common occurrence in research robots, the robots' movement might be aborted by reflex errors. This is mainly due to sudden and non-smooth movements. To avoid these reflexes, we gradually increase the gain (from zero to a final constant value) at the beginning of the servoing tasks. We also saturate the velocities (translational and rotational) sent to the robots.

\begin{figure*}[!t]\centering
    \includegraphics[trim={0 10pt 0 10pt},width=.99\linewidth]{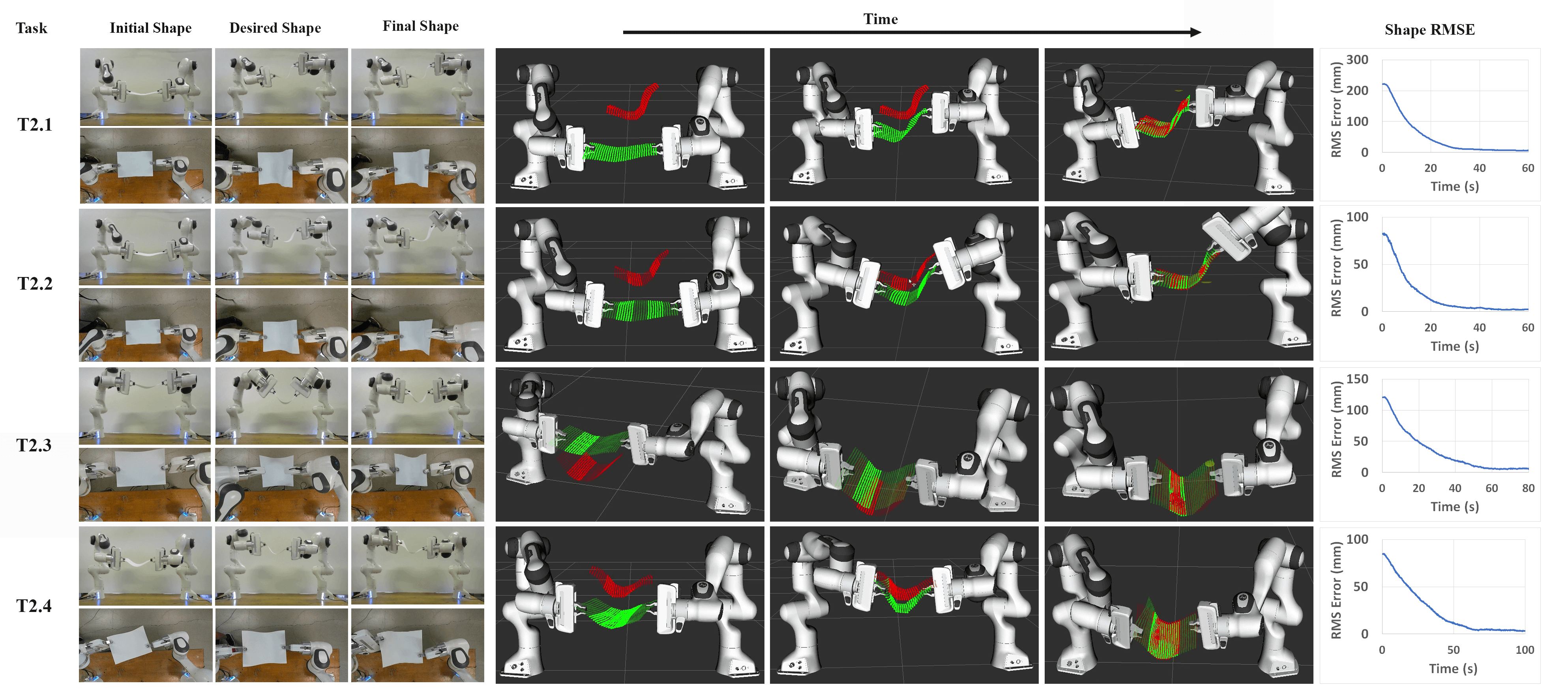} 
    \caption{Tasks with a thin-shell object: an A4 paper. Top row: task with full shape servoing toward a desired shape with large deformation. Second row: task with partial shape servoing. Two separated regions (one-fifth of the paper) are servoed. Third row: task with partial shape servoing. One fourth of the A4 paper is servoed. Only translational velocities of the robots are updated. Fourth row: task with partial shape servoing with a slightly thicker A4 paper. Two-fourths middle part of the object is servoed. The grippers are displaced while setting the initial shape.}
    \label{fig:thinshell_paper}
\end{figure*}

\begin{figure*}[!t]\centering
    \includegraphics[trim={0 10pt 0 10pt},width=.99\linewidth]{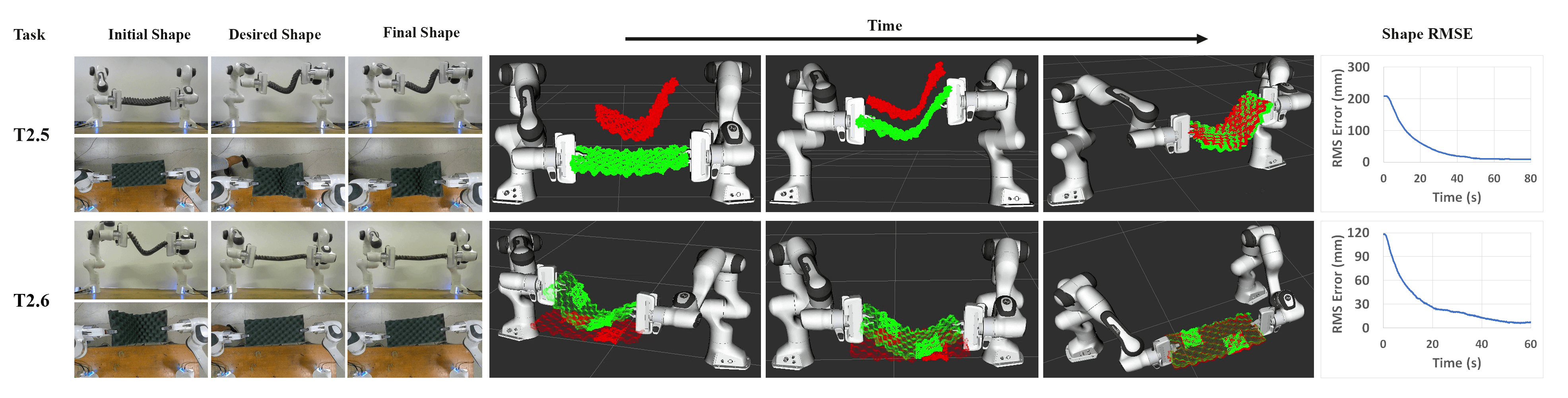} 
    \caption{Tasks with a thin-shell object: a convoluted foam. Top row: task with full shape servoing. Bottom row: task with partial shape servoing. Two small separated regions of the foam are servoed.}
    \label{fig:thinshell_packingfoam}
\end{figure*}

In total, we define thirteen tasks, each with specific features and challenges. Table \ref{tab:experiments} presents the main parameters of each task. We categorize the tasks based on the general form of the object under manipulation.
The tasks' results are presented in figures \ref{fig:experiments_linear} to \ref{fig:volumetric_octagonal}.
In each figure, the elements corresponding to the current and desired shapes are visualized with green and red colors, respectively.
Furthermore, the sections of the object or lattice belonging to the servoed regions are indicated in brighter colors while the ones belonging to the free regions (which are present in partial shape servoing) are indicated in darker colors. 
Finally, for each task, an RMSE graph (RMS of $\mathbf{e}_s$) illustrates the servoing error during the task.
We set the control gain $k_{p}$ as 0.1 in full shape servoing tasks and 0.05 in partial shape servoing tasks. 
We tune these gain values empirically. They allow us to obtain good performance while avoiding reflex errors that cause the robots to stop moving.
In the following, we explain the tasks in more detail.

\subsection{Linear objects} \label{subsec:Linear_objects}
For the experiments with linear objects, we use an electric cable with the length of 70\,cm.
As for creating $\mathbf{p}^u$, we form a cylinder with the same dimension as the cable. However, instead of considering the whole point cloud on the surface of this cylinder, we consider merely half of the points, i.e., a semi-cylinder. This is done to prevent the inferred object shape and the lattice from rotating around the longitudinal axis during tracking due to axial symmetry.
We form a $15\times 3\times 3$ size lattice around $\mathbf{p}^u$ in a way that the lattice direction with 15 nodes is in line with the length of the cable.
The cable is manipulated from its two ends by the two robots.
We define three tasks with this cable: two in-plane, and one out-of-plane.
We put a board between the two robots on which we lay the cable. 
This board serves two purposes; first, the in-plane manipulations take place on the surface of this board, and second, we employ the difference in colors of the board and the cable to filter out unwanted captured point cloud coming from the board.
The results of the tasks can be found in figures \ref{fig:experiments_linear} and \ref{fig:experiments_linear_outofplane}. In figure \ref{fig:experiments_linear}, the plotted points represent the mean coordinates of the lattice nodes belonging to the cross-sections of the lattice along the cable length.
In the following, we explain the tasks in more detail. 

\begin{itemize}
\item 
\textit{T1.1}. In this task, we fully servo the cable toward a shape with a large in-plane deformation. 
In order to make sure that the deformation remains in-plane, while setting the initial and the desired shapes, we keep the robots' grippers parallel to the board at a slight distance above the surface of the board. Furthermore, during servoing, we only send the velocity elements to the robots that keep the robot in the same distance with respect to the board, i.e., two translational velocities parallel to the board and the rotational velocity perpendicular to the board.

\item 
\textit{T1.2}. This task is similar to T1.1. The main difference is that we partially servo the cable. To this end, we divide the cable into 5 sections along its length and select the three middle sections as the servoed region of the object. We then select the lattice nodes encapsulating this region as the servoed nodes of the lattice.

\item 
\textit{T1.3}. This experiment aims to servo the cable through an out-of-plane deformation. To this end, we remove the constraints regarding keeping the grippers' relative pose with respect to the board. This is done by setting the grippers higher in comparison to the board's surface while defining the desired shape. This can be observed in figure \ref{fig:experiments_linear_outofplane}. We also send the full translational and rotational velocities to the robots. 
\end{itemize}

\subsection{Thin-shell objects} \label{subsec:thin-shell_objects}
The next experiments are conducted with thin-shell objects. Our objects of interest are a blank A4 paper and a convoluted foam.
For both objects, we form a $8\times 8\times 3$ size lattice where the $8\times 8$ side is aligned with the surface of the objects and the direction with 3 nodes is in line with the width of the objects.
We start with the paper.
We define four tasks which are explained in the following. Figure \ref{fig:thinshell_paper} presents the results of these tasks.

\begin{itemize}
\item 
\textit{T2.1}. In this task, we fully servo the paper toward a desired shape with large deformation. 

\item 
\textit{T2.2}. This task aims to do partial servoing with the paper. To this end, similarly to T1.2, we divide the paper into five sections along its longer dimension. We, then, select the lattice nodes encapsulating the second and the fourth sections of the paper as the servoed lattice nodes.  

\item 
\textit{T2.3}. This task is similar to T2.2. There are, however, two main differences: first, the servoed region is smaller, i.e., one-fourth of the object, and second, we merely apply translation to the robots' grippers without any rotation. The latter is done by updating only the translational velocities of the robots.

\item 
\textit{T2.4}. In this task, we do partial shape servoing with two-fourth middle region of the paper as the servoed region. The important change in this task is that we significantly displace the grippers on the paper while setting the initial shape (each in the opposite corner of the paper) in comparison to the desired shape (both in the middle of the paper). We also update the grasped lattice nodes after setting the initial shape. This task is performed with a slightly thicker A4 paper so that the corners of the paper do not entirely loosen due to the large distance from the grippers.

\end{itemize}

\begin{figure*}[!t]\centering
    \includegraphics[trim={0 10pt 0 10pt},width=.99\linewidth]{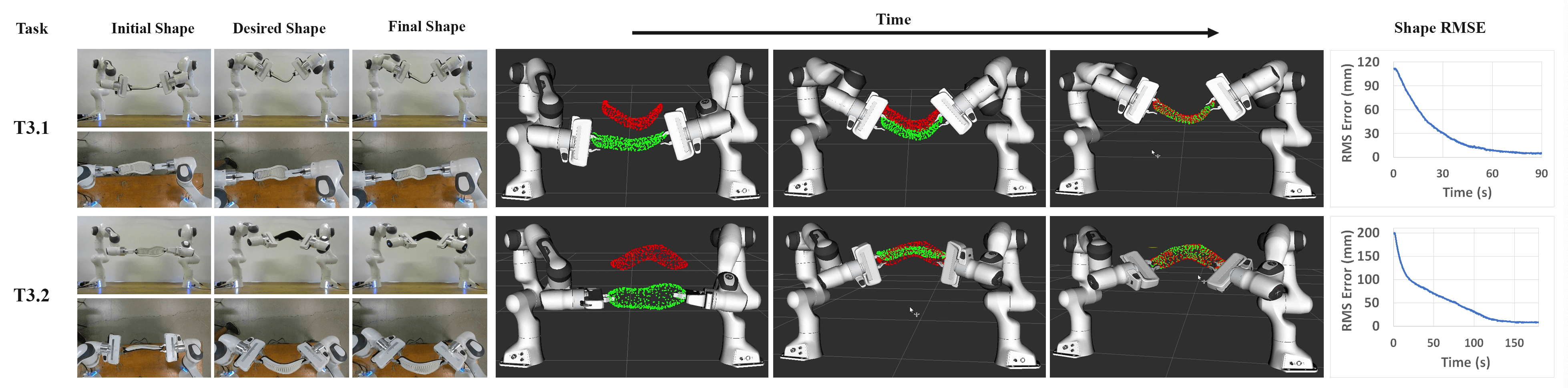} 
    \caption{Tasks with a volumetric object: a bulky shoe sole. Top row: task with large deformation. Bottom row: task with a desired shape with a considerably different view of the object with respect to the initial shape. In this task, the object is rotated and flipped by the shape servoing approach.}
    \label{fig:volumetric_shoesole}
\end{figure*}

\begin{figure*}[!t]\centering
    \includegraphics[trim={0 10pt 0 10pt},width=.99\linewidth]{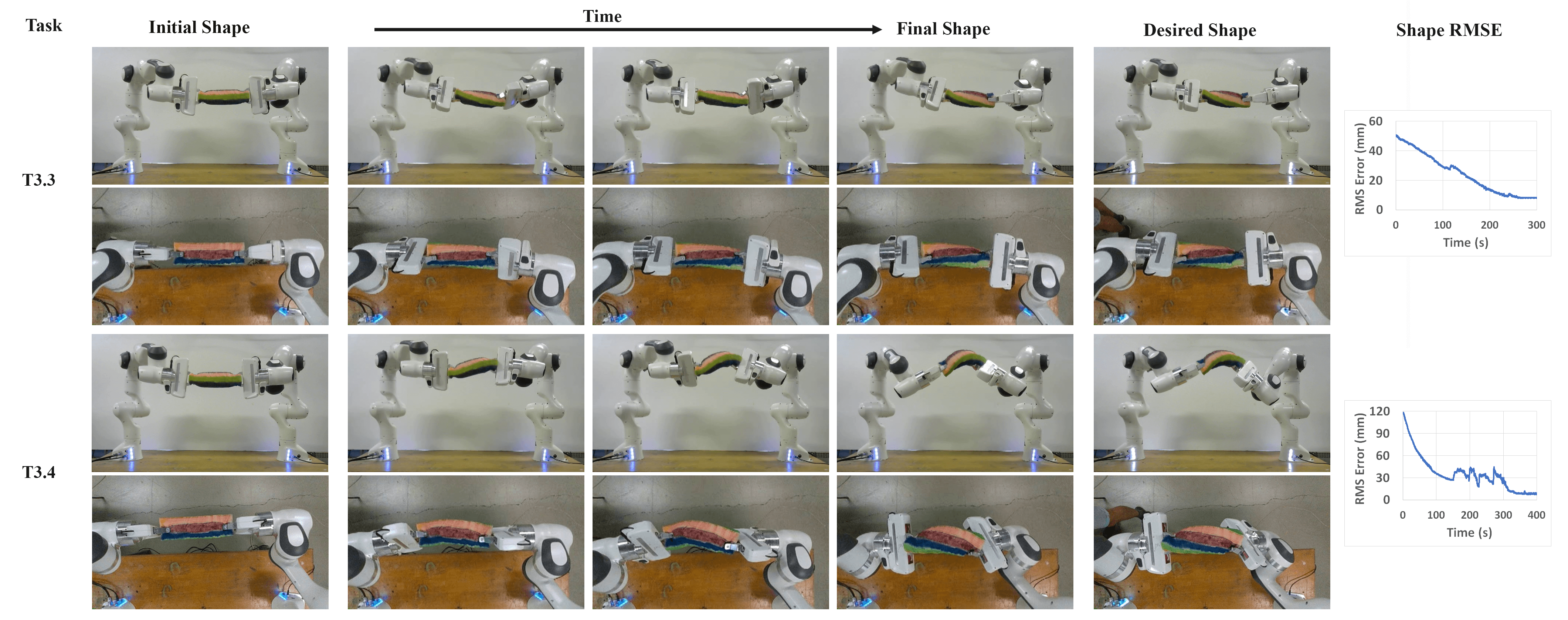} 
    \caption{Tasks with a volumetric object: a foam octagonal cylinder. Top row: task with merely twist. Bottom row: task with twist and bending deformation at the same time.}
    \label{fig:volumetric_octagonal}
\end{figure*}

The next series of experiments are with a convoluted foam that is widely used in packaging industry.
The goal is to demonstrate that the surface of the thin-shell object should not necessarily be flat.
This is thanks to the generality that using the lattice brings to our approach as we can track and servo objects of any geometry.
Another point is that as the convoluted foam is thin and has a low stiffness, it does not follow the rotations of the grippers. We, thus, similarly to T2.3, update merely the translational velocities of the robots. 
We define two tasks with the convoluted foam which are described in the following. 
Figure \ref{fig:thinshell_packingfoam} presents the results of these tasks. 

\begin{itemize}
\item 
\textit{T2.5}. In this task, we conduct full shape servoing toward a desired shape with a large deformation with respect to the initial shape.

\item 
\textit{T2.6}. This task aims to conduct a partial shape servoing with two small separated servoed regions of the convoluted foam. To this end, we initialy select the same servoed regions as in T2.2 and then inversely halve each region relative to the center-line of the convoluted foam. In contrast to the previous tasks, we define the servoed regions' desired shapes in a way that makes the convoluted foam undeformed at the end of the task.
\end{itemize}

\subsection{Volumetric objects} \label{subsec:volumetric_objects}
The final set of experiments is carried out with two volumetric objects: a bulky shoe sole, and a foam octagonal cylinder.
For both objects, we form an $8\times 4\times 4$ size lattice where the direction with eight nodes is in line with the longer direction of the objects. 
We start with the shoe sole. 
We define two tasks with the shoe sole which are explained in the following. 
Figure \ref{fig:volumetric_shoesole} presents the results of these tasks.

\begin{itemize}
\item 
\textit{T3.1}. In this task, we fully servo the shoe sole toward a desired shape with a large deformation. 

\item 
\textit{T3.2}. This task is similar to T3.1. The main difference is that the desired shape is defined with a severe rotation with respect to the initial shape. This can be observed in figure \ref{fig:volumetric_shoesole}. In fact, the shoe sole should be rotated and flipped during the servoing. 
Consequently, the part of the shoe sole that is visible in the desired shape is considerably different from the one in the initial shape. 
\end{itemize}

As the last set of experiments, we define two tasks with the foam octagonal cylinder. 
These tasks include twisting the foam. Hence, in order to ensure that the foam follows the rotations of the grippers, we select the foam to be relatively dense. 
This applies an intense rotational force to the robots during the servoing.
We, thus, considerably decrease the saturation values of the grippers' velocities to avoid reflex errors in the robots.
We also paint the foam lengthwise to better illustrate the applied twist. 
Figure \ref{fig:volumetric_octagonal} presents the results of these tasks.
The defined tasks are described in the following.

\begin{itemize}
\item 
\textit{T3.3}. In this task, the desired shape is set by applying a pure twist to the foam. 

\item 
\textit{T3.4}. In this task, we apply twist and bending deformation at the same time. 
\end{itemize}

\subsection{Results and discussion} \label{subsec:results-and-discussion}
As shown through various experiments, our proposed unified tracking-servoing approach is able to track different forms of the objects and fully and partially servo them toward largely deformed desired shapes. 
The servoing error graphs in the rightmost side of the figures \ref{fig:experiments_linear} to \ref{fig:volumetric_octagonal} verify the efficiency of our approach in different scenarios. 
The variety of objects' materials employed in these experiments confirms the robustness of our proposed approach in dealing with many of the elastic deformable objects around us without having a knowledge of their mechanical parameters.
Using a 3D lattice makes it possible to use the same tracking and servoing approach for an object with any form.
Furthermore, as shown, our approach can handle tasks with partial shape servoing with one or multiple small servoed regions. Defining servoed regions is quite straightforward and is done by just specifying the corresponding servoed lattice nodes encapsulating the object's servoed regions. 
Another point that should be noted here is that our approach provides full control over the deformation of the object in 3D space; i.e., the object can be simultaneously deformed, rotated and translated.
This can be particularly observed in T3.2 where the visible side of the shoe sole in its desired shape is totally different in comparison to the one in its initial shape.
Applying twist and bending deformation in T3.3 and T3.4 is another manifestation of this full control over the object's deformation in 3D space. 
To the best of our knowledge, no approach in the literature possesses this feature.
The servoing can be even performed in the existence of some noise in the tracking as can be seen in T3.4.
Note that the perturbations seen in the middle part of this task are due to tracking noise, and not related to the servoing performance. Next, we discuss several more specific aspects of the implementation and performance of our proposed approach. Note that various results from tests in simulation to further illustrate some of these aspects are available on our project website. 

\subsubsection{Unreachable shapes} There might be cases where the desired shape is unreachable. This can be due to three main reasons: \textit{(i)} the intrinsic properties of the desired shape, i.e., it is not reachable by the current shape of the object from the current grasping points, \textit{(ii)} the existing movement constraints in the robots, e.g, the desired shape is defined out of the task space of the robots, and \textit{(iii)} manipulation constraints, i.e., the interaction between the grippers and the elastic deformable object is in a way that one or several degrees of freedom are practically lost. An example of the latter is when the object is too soft to follow rotations (T2.5 and T2.6). 
When dealing with these cases, our proposed servoing pipeline
drives the object toward a shape with a small residual error that corresponds with a global minimum with a non-zero value. This can be observed in T2.3, T2.5, and T2.6.

\subsubsection{Rigid motion and time horizon}
Our servoing pipeline does not make a distinction between rigid and nonrigid components of the shape servoing error. This is the common approach in the state-of-the-art \cite{Duenser2018,Navarro2018,berenson2013manipulation}. Our extensive experiments are carried out in scenarios close to those of interest in real-world industrial applications, and they involve simultaneous rigid and nonrigid object motions. As our experimental results demonstrate, our servoing pipeline performs with efficiency and accuracy in these scenarios. In cases with a very large rigid motion of the object between the initial and desired shapes, it may be interesting to distinguish rigid from nonrigid components of the shape servoing error. This could enable finer control over the robotic arm motions and the evolution of the object's shape. In this respect, we see no hurdles to combining our approach with a specific module for rigid-body motion handling (which is a well-studied problem). 
In addition, we conducted preliminary simulations that suggest that our Jacobian estimation could be applicable for a certain time horizon in an MPC-style controller; especially, for moderate deformations. These aspects, which exceed the scope of our paper, are interesting directions for future work.

\subsubsection{Computational cost}
Regarding the execution speed of our approach, during the experiments, we reached 20-30\,FPS for the whole process (tracking and servoing) without any parallelization using only CPU. This range is the same as in \cite{shetab2022rigid}. The main difference is that our approach handles both tracking and servoing at the same time while in \cite{shetab2022rigid}, the execution speed is reported only for servoing, and tracking is performed by a separate algorithm. 
One point that should be noted here is that in calculating the analytical Jacobian, the inversion of matrices in (\ref{eq:J_sg}) would be costly when applied to a very large mesh. 
In our approach, however, this does not cause a problem as we calculate the deformation Jacobian for the lattice whose shape has a much smaller size (in terms of number of nodes) than the object.

\subsubsection{Non-convex object geometries}\label{sec:nonconvex}
In this work, we only consider objects with convex geometry, and we form a uniform lattice over them. If the object has a non-convex geometry, using a uniform lattice can result in unnecessary geometric constraints between different parts of the object that are not directly connected. For instance, in the case of a doll toy, such constraints may arise between the hands and legs, leading to unrealistic deformations where the movement of one hand affects the position of a leg. To overcome this challenge, a method proposed in \cite{zollhofer2012gpu} can be used, where unnecessary links in the lattice are disregarded. This can be achieved by creating a uniform lattice around the non-convex object and then removing all tetrahedral cells and their corresponding lattice nodes that lie entirely outside the input geometry. We note that our approach can be applied to the modified lattice without any modifications.

\subsubsection{Correspondences between the current and desired shapes of the lattice} Our servoing pipeline is applied to the lattice, and requires correspondences between the servoed nodes of the current and desired shapes of the lattice. Next, we explain in more detail how the experimental process described in Sect. \ref{subsec:experimental_setup} gives us these correspondences. As the first step of this process, we launch our tracking pipeline. Then, we manually set the desired shape. After that, we manually set the initial shape, i.e., the current shape at time zero. And finally, we launch the servoing pipeline. Note that the tracking pipeline is running continuously, i.e., tracking the object and lattice continuously, throughout all these steps. Hence, we get the correspondences between all the lattice nodes, including the set of servoed nodes, of the current and desired shapes. This type of approach is practical and we have used it previously \cite{shetab2022rigid}. One of its advantages is that it employs short-baseline registration between each two consecutive frames during tracking (see Sect. \ref{sec:finding_corresponding_points}), instead of wide-baseline registration between the current and desired shapes, which can be more challenging.

\subsubsection{Task-specific adjustments}
To conclude, we discuss the implications of the adjustments we made in several experiments. We restricted the robots' action space in several tasks: in T1.1, T1.2 we wanted to mimic a planar manipulation scenario, so as to have a more diverse set of experiments; but this restriction is not needed, as shown by our full-action-space experiments with 3D gripper motion. Similarly, in T2.3, T2.5 and T2.6 we did not use rotation of the grippers. We did this to illustrate the behavior in cases where this limitation exists, but this is not a limitation of our approach. We also chose the initial and desired shapes in our tests to illustrate a diversity of situations, including large deformations and cases of interest in industrial application (e.g., folding or unfolding). In addition, it is common and reasonable to select a lattice that represents the object's deformation suitably (as discussed in sections \ref{sec:Problem_formulation}, \ref{sec:nonconvex}), and that also facilitates tracking of the object (e.g., the lattice for the elastic cable in our experiments). Clearly, in practice, every particular scenario has its own particular features that need to be considered. However, neither this fact nor the adjustments described above detract from the generality of our approach.

\begin{figure}[!t]\centering
    \includegraphics[trim={0 10pt 0 10pt},width=.99\linewidth]{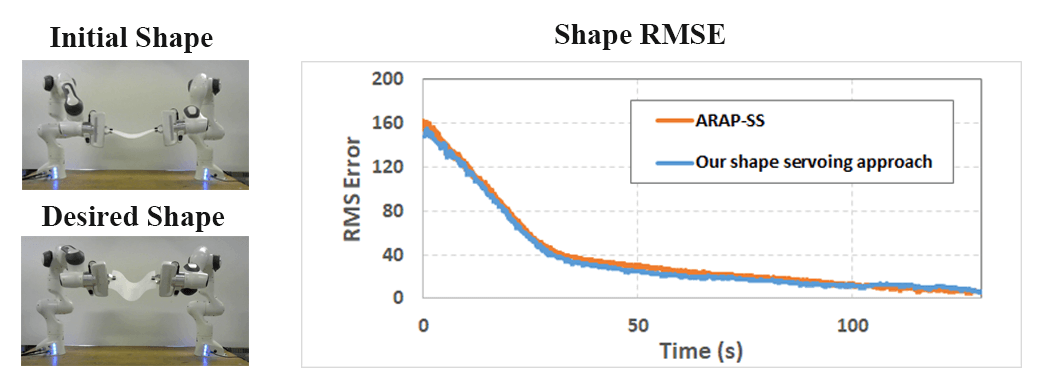} 
    \caption{Comparison between ARAP-SS and our proposed shape servoing pipeline. Left: initial and desired shapes which are identical for both approaches. Right: shape servoing RMS error over time belonging to both approaches.}
    \label{fig:comparison}
\end{figure}

\subsection{Comparison with state-of-the-art} \label{subsec:comparison_with_ stateoftheart}
In this section, we compare our shape servoing pipeline with our previously presented ARAP shape servoing (ARAP-SS) scheme \cite{shetab2022rigid} through an experiment. ARAP-SS considered thin-shell objects, and it did not provide unified tracking and servoing as we do here. We design two identical tasks for the two approaches with the thick A4 paper from T2.4. The initial and the desired shapes are considered the same for the two tasks as can be seen in figure \ref{fig:comparison}. This is done by storing the robots' configuration for both initial and desired shapes and keeping the grippers' poses on the object unchanged in the two tasks. 

ARAP-SS was designed for thin-shell objects, as it was based on a surface (not volumetric) deformation model. Thus, we use only the nodes and interconnections on the outer surface of the lattice as the template to ARAP-SS. We thus servo only these lattice nodes from the initial to the final shape.  
In order to have a fair comparison, we do a partial shape servoing with the same outer lattice nodes with our proposed servoing pipeline. 
We also use the same gripped nodes and control gains in both approaches. 
Similarly to the previous tasks, we saturate the translational and rotational velocities sent to the robots. 
In the right-hand side of figure \ref{fig:comparison}, the shape servoing errors of the two approaches are compared. 
As seen, no significant difference can be observed between these graphs.
This validates the precision of our servoing approach
in comparison to the precision (which is state-of-the-art) of ARAP-SS.
This precision, along with other features of our proposed approach, including the analytical expression for the Jacobian, having full control over the object's deformation in 3D space, and scalability, privilege our approach with respect to other existing approaches.
 

\begin{figure*}[!t]\centering
    \includegraphics[trim={0 10pt 0 10pt},width=.99\linewidth]{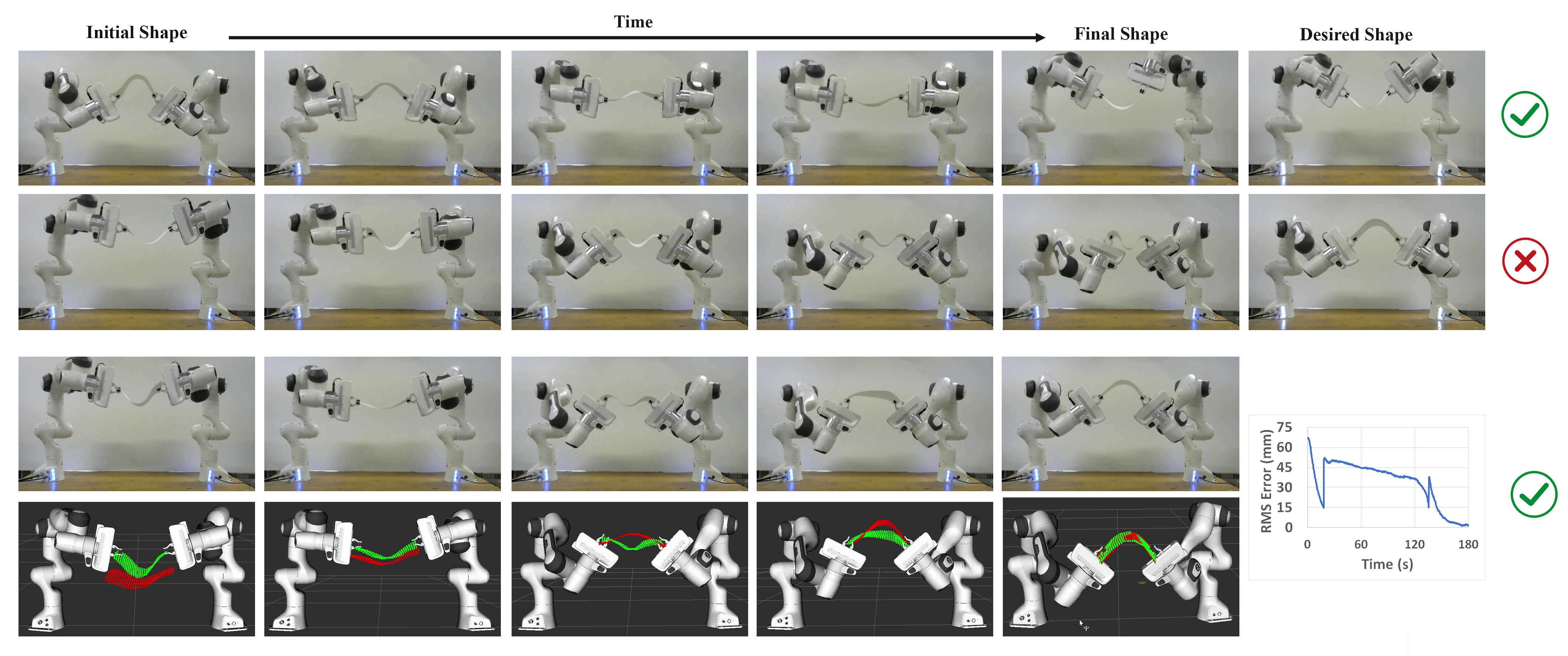} 
    \caption{ Tasks for driving a thick A4 paper through its singular shape, i.e., a flat shape. First row: successful task, deforming the paper from an upward curved shape to a downward curved shape. Second row: unsuccessful task, deforming the paper from a downward curved shape to an upward curved shape (deforming against gravity). Two last rows: successful task, defining two intermediary desired shapes for performing the failed task of the second row.}
    \label{fig:singularity}
\end{figure*}

\subsection{Failed cases} \label{subsec:failed_cases}
In this section, we present a failed case of our approach. 
Our failed case concerns driving the object through its singular shape, i.e., deforming it from an upward curved to a downward curved shape or vice versa.
In general, this is a complicated task as it requires certain techniques that might differ from one object to another. 
We show this difficulty by defining tasks with the thick A4 paper from T2.4.
These tasks can be observed in figure \ref{fig:singularity}.
We start with a task in which the initial shape is upward curved, and the desired shape is downward curved. This is shown in the first row of figure \ref{fig:singularity}. As seen, the shape servoing pipeline can successfully drive the object through the flat shape, i.e., the singular shape. It should be noted that, for this task, the direction of gravity is favorable throughout the servoing. 
In order to make the task more challenging, this time, we try to start from a downward curved  shape and drive the object to an upward curved  shape. 
This is shown in the second row of figure \ref{fig:singularity}.
As seen, despite the severe rotations applied to the paper by the robots, the paper cannot pass through the flat shape. This continues until the robots reach their rotational limits. 
We show that it is possible to solve this problem using a planning strategy. In particular, this can be done by firstly unwrapping the paper to a certain extent and then applying the required rotation to drive the paper through its singular shape. 
We applied this approach by defining two intermediary desired shapes for the paper: one nearly unwrapped downward curved  shape, and one nearly unwrapped upward curved  shape. We switch from one desired shape to the next one when the shape servoing RMS error becomes smaller than 15\,mm. The last two rows of figure \ref{fig:singularity} present this process.
As seen, using this solution, the task can successfully be carried out. 

\section{Conclusion}
In this paper, we present a general tracking-servoing approach that is capable of deforming elastic objects toward a desired shape. Our approach has \color{black}full control over the deformation of elastic objects of any form (linear, thin-shell, volumetric) and any geometry in 3D space. \color{black}
Next, we mention several limitations of our approach. For objects with low stiffness (like a cloth), it might be required to incorporate gravity which is not considered in our approach. \color{black}We cannot servo any arbitrarily complex deformation of an object, but only those deformations that can be captured by the lattice. \color{black} Besides, reliably driving the object through a singular shape requires additional techniques. Furthermore, handling contact with the environment is not included in our approach. Finally, as we used our Jacobian with a Cartesian velocity controller, there is no constraint for keeping the robots in comfortable configurations throughout the task. This might lead to the failure of the task. A direction for future work is to employ additional constraints for alleviating the last limitation. Considering contact with static objects in the scene or incorporating a planning algorithm to define feasible intermediary shape targets in particularly challenging scenarios are two other interesting future directions. Furthermore, as we use a 3D lattice for any object, it is possible to transfer deformation from one lattice to another and consequently from one object to another. This feature can be useful in motion transfer applications concerning \color{black}elastic \color{black} deformable objects. We also consider using the idea of employing a lattice for improving generalization in reinforcement learning for elastic deformable objects of any shape.

 \balance
\bibliographystyle{IEEEtran}
\bibliography{Text}

\vspace{11pt}
\vspace{-33pt}
\begin{IEEEbiography}[{\includegraphics[width=1in,height=1.25in,clip,keepaspectratio]{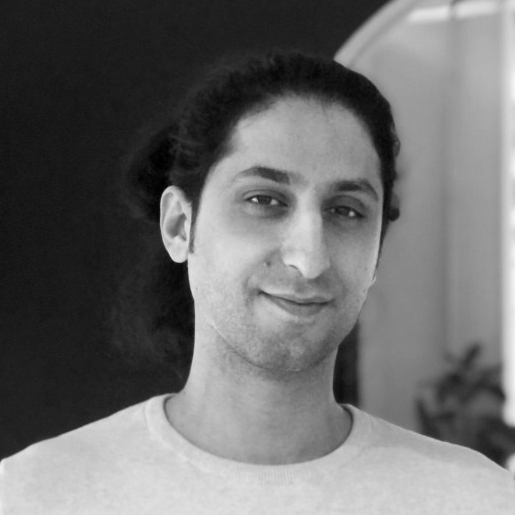}}]{Mohammadreza Shetab-Bushehri}
received his Ph.D. in Electrical, Electronic, and Engineering Systems from the Université de Clermont Auvergne in 2023. Presently, he holds the position of Research Fellow at the Ecole Centrale de Lyon in France. His current research interests include the perception and manipulation of deformable objects and the development of multi-task robotic agents.
\end{IEEEbiography}

\begin{IEEEbiography}[{\includegraphics[width=1in,height=1.25in,clip,keepaspectratio]{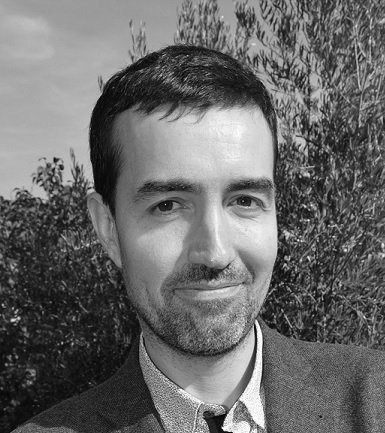}}]{Miguel Aranda}
 received the Ph.D. degree in systems engineering and computer science from the Universidad de Zaragoza, Spain, in 2015. He has held postdoctoral research positions within the Institut Pascal laboratory (UMR 6602) in Clermont-Ferrand, France. Currently, he is a Research Fellow with the Instituto de Investigación en Ingeniería de Aragón (I3A) at the Universidad de Zaragoza. His current research interests include multiagent control and robotic manipulation.

\end{IEEEbiography}
\begin{IEEEbiography}[{\includegraphics[width=1in,height=1.25in,clip,keepaspectratio]{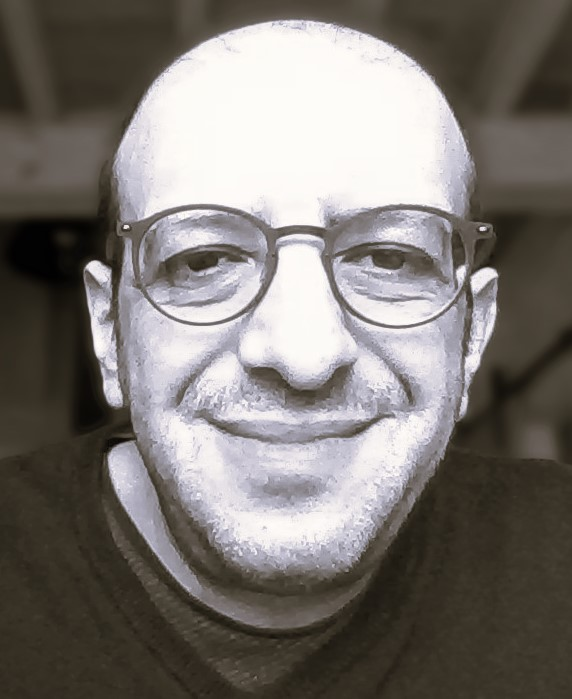}}]{Youcef Mezouar}
received the Ph.D. degrees in automation and computer science from the University of Rennes 1, France in 2001. He obtained the Habilitation Degree (HDR - Habilitation à Diriger des Recherches) from Université Blaise Pascal,  Clermont-Ferrand, France, in 2009. He spent one year as Postdoctoral Associate in the Robotic Lab of the Computer Science Department of Columbia University, New York. He was assistant professor from 2002 to 2011 in the Physics Department of Blaise Pascal University, Clermont-Ferrand, France. He is a Full Professor at Clermont Auvergne INP-SIGMA’Clermont  since 2012.  His research interests include sensor-based control and computer vision.
\end{IEEEbiography}

\begin{IEEEbiography}[{\includegraphics[width=1in,height=1.25in,clip,keepaspectratio]{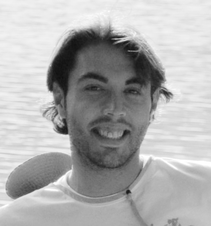}}]{Erol \"{O}zg\"{u}r}
 received the Ph.D. degree in Robotics and Vision from the Université Blaise Pascal, France (2012).
Later, he was a postdoctoral fellow in Institut Pascal-UBP/CNRS/IFMA, France (2012-2015).
Afterward, he joined as an assistant professor at the Université d’Auvergne, France (2015-2018).
Now, he is an assistant professor in Clermont Auvergne INP - Sigma Clermont (2018).
His research interests are vision-based robot control and computer vision.
\end{IEEEbiography}

\end{document}